\pdfoutput=1
\documentclass[11pt]{article}

\usepackage[final]{acl}      

\usepackage{times}
\usepackage{latexsym}

\usepackage[T1]{fontenc}
\usepackage[utf8]{inputenc}

\usepackage{microtype}

\newcommand{\alg}{\textbf{\texttt{APEX}}\xspace}
\newcommand{\cmark}{\ding{51}}%
\newcommand{\xmark}{\ding{55}}%
\usepackage{mathtools}

\usepackage{microtype}

\usepackage{graphicx, subcaption, booktabs}
\usepackage{multicol, multirow}

\usepackage{amsmath, amssymb, amsthm}
\usepackage{algorithm, algorithmic} 

\usepackage{tabulary}
\usepackage{wrapfig, makecell}
\usepackage{enumitem}
\usepackage{pifont}
\usepackage{colortbl}
\usepackage{xspace}
\usepackage{kotex}
\usepackage{resizegather}
\usepackage{xcolor}

\title{Towards Difficulty-Agnostic Efficient Transfer Learning for Vision-Language Models}

\newtheorem{definition}{Definition}
\newtheorem{observation}{Observation}

\author{Yongjin Yang\thanks{\, equal contribution} \quad Jongwoo Ko\footnotemark[1] \quad Se-Young Yun \\
  KAIST AI \\
\texttt{\{dyyjkd, jongwoo.ko, yunseyoung\}@kaist.ac.kr}
  }

\begin{document}
\maketitle
\vspace{-15pt}
\begin{abstract}
Vision-language models (VLMs) like CLIP have demonstrated remarkable applicability across a variety of downstream tasks, including zero-shot image classification. 
Recently, the use of prompts or adapters for efficient transfer learning (ETL) has gained significant attention for effectively adapting to downstream tasks. 
However, previous studies have overlooked the challenge of varying transfer difficulty of downstream tasks. In this paper, we empirically analyze how each ETL method behaves with respect to transfer difficulty. 
Our observations indicate that utilizing vision prompts and text adapters is crucial for adaptability and generalizability in domains with high difficulty.
Also, by applying an adaptive ensemble approach that integrates task-adapted VLMs with pre-trained VLMs and strategically leverages more general knowledge in low-difficulty and less in high-difficulty domains, we consistently enhance performance across both types of domains.
Based on these observations, we propose an adaptive ensemble method that combines visual prompts and text adapters with pre-trained VLMs, tailored by transfer difficulty, to achieve optimal performance for any target domain. 
Upon experimenting with extensive benchmarks, our method consistently outperforms all baselines, particularly on unseen tasks, demonstrating its effectiveness.
\end{abstract}
\section{Introduction}

Vision-language models~(VLMs), such as CLIP~\cite{radford2021learning} and ALIGN~\cite{jia2021scaling}, have demonstrated remarkable applicability across various downstream tasks such as image classification. 
A distinctive feature of these VLMs for image classification is their ability to classify unseen classes that have not been encountered during pre-training through zero-shot inference, which is not possible to traditional vision models.

The primary challenge of VLMs for downstream tasks is to excel in classifying both seen and unseen class sets. In the context of VLM classification tasks, the ability to accurately classify seen class sets is termed \textit{adaptability}, while the capability to extend this proficiency to unseen class sets is referred to as \textit{generalizability}. To boost these abilities, recent research has introduced efficient transfer learning~(ETL) methods to fine-tune VLMs. One strategy involves the use of soft prompt tuning~\cite{zhou2022learning, zhou2022conditional, khattakMaPLe, khattak2023self}. Another research direction involves adapter-style tuning \cite{gao2023clip, zhang2022tip, zhu2023not} either by adjusting specific parameters or employing cache-based techniques. 
These approaches empower VLMs to swiftly adapt to new tasks using only a few samples (i.e. few-shot image classification task).

However, previous approaches do not consider a significant factor for adapting to downstream tasks: varying transfer difficulty~\cite{yu2023task}. This refers to the challenge of adapting pre-trained VLMs according to the target domain. For instance, transferring pre-trained VLMs to specific fine-grained domains, such as FGVC Aircraft, is more challenging than transferring to general coarse-grained domains. In a real-world scenario, it is hard to predict the specific target task and domain that will emerge. Therefore, without investigating how each type of ETL behaves in response to different levels of transfer difficulty and applying an adaptive method based on this investigation, the result for each target domain can be suboptimal. Some works manually train models differently for each dataset~\cite{gao2023clip, zhang2022tip}, but this approach is not feasible in real-world scenarios as prior knowledge for the target task is not given.

\begin{figure*}[t]
    \centering
    \includegraphics[width=1.0\linewidth]{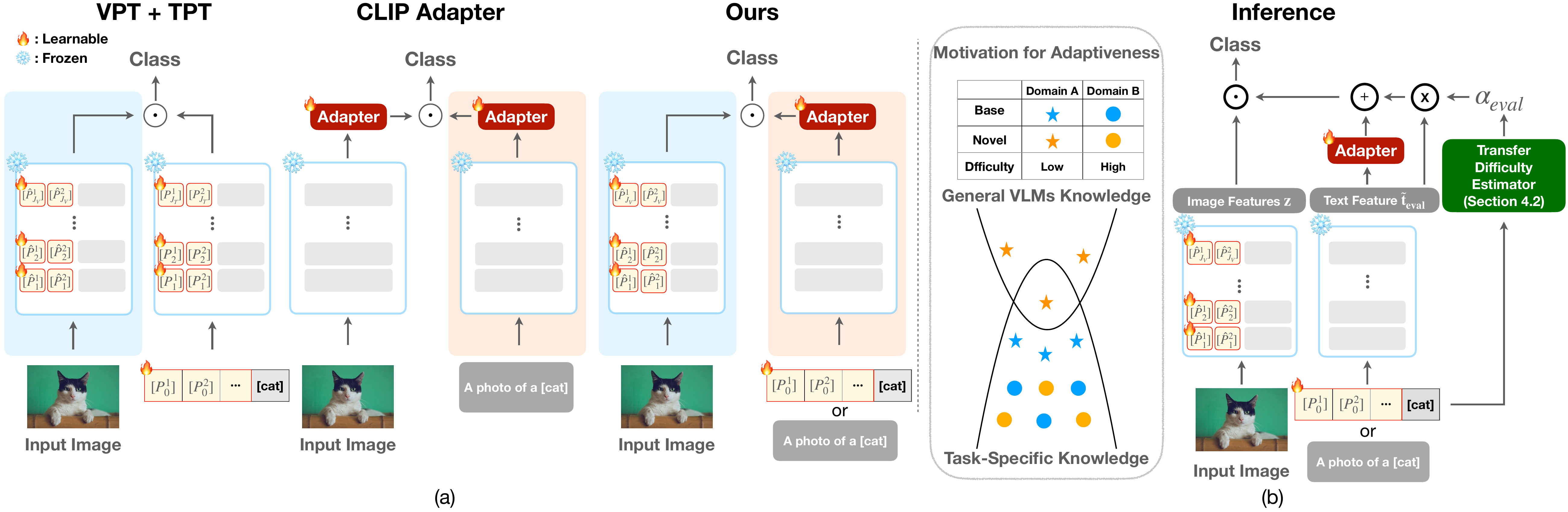}
    \caption{
    Overview of \alg compared to the conventional ETL methods. \alg exhibits two key differences: 
    \textcolor{blue}{\textbf{(a)}:} \textit{Firstly}, \alg integrates prompt tuning for the visual encoder and a linear adapter for the text encoder, each tailored to the specific properties of their respective modalities, which performs better on high-difficulty domains.
    \textcolor{blue}{\textbf{(b)}:} \textit{Secondly}, \alg integrates an adaptive coefficient within the text encoder to strategically balance pre-adapter and post-adapter features to properly combine task-specific knowledge and general VLMs knowledge based on transfer difficulty. A detailed explanation, including notations and the algorithm, can be found in Section~\ref{sec:method} and Appendix~\ref{appendix:notation_algorithm}.
    }
    \label{fig:intro_figure}
\end{figure*}

To overcome these limitations and apply an adaptive method for tuning adapters and prompts for downstream tasks, we empirically investigate the characteristics of applying different tuning methods for each modality on multiple domains with varying transfer difficulty, revealing four key findings.
\textit{Firstly}, we find that visual prompt tuning (VPT) generalizes better to unseen classes compared to text prompt tuning (TPT) in cases of high-difficulty domains, as TPT tends to overfit on base classes for these domains.~($\triangleright$ Obs.~\ref{obs:obs_1}).
This occurs because, in high-difficulty domains, the class separability of visual features from a visual encoder is low, causing TPT to overly adapt in classifying these challenging features~($\triangleright$ Obs.~\ref{obs:obs_2}). 
\textit{Moreover}, text adapter (TA) can significantly boost the adaptability of VPT, resulting in high adaptability and generalizability, especially for highly difficult domains~($\triangleright$ Obs.~\ref{obs:obs_3}). However, fine-tuning with adapters could compromise generalizability in easier domains. Our \textit{last} observation is that combining pre- and post-adapter features to leverage pre-trained VLMs knowledge can address this concern with a proper balance between them. For instance, using more pre-adapter features can maintain generalizability in easier domains. The ideal balance depends on the domain's difficulty, highlighting the need to adjust the ensemble coefficient accordingly~($\triangleright$ Obs.~\ref{obs:obs_4}).

Based on our observations, we present a \alg~(text \textbf{\underline{A}}dapter, visual \underline{\textbf{P}}rompt, and adaptive \textbf{\underline{E}}nsemble for cross(\textbf{\underline{X}})-modality) that utilizes an adaptive ensemble with VPT and TA. Specifically, we use the combination of VPT and TA, which have shown high generalizability and adaptability for high-difficulty domains, as shown in Obs.~\ref{obs:obs_1}-\ref{obs:obs_3} (Fig.~\ref{fig:intro_figure}(a)). Also, motivated by Obs.~\ref{obs:obs_4}, we employ an adaptive ensemble approach that determines the optimal ensemble coefficient for each domain by using the distances to learned classes in pre-trained VLMs to estimate transfer difficulty (Fig.~\ref{fig:intro_figure}(b)).
This adaptive ensemble controls the level of adaptation, by primarily utilizing task-specific knowledge with adapted VLMs for high-difficulty domains but leveraging general knowledge for low-difficulty domains, as pre-trained VLMs already possess sufficient ability and prevent an overfitting from excessive adaptation.
With this, our method acts as a difficulty-agnostic solution, enabling the model to effectively adapt to all target domains regardless of transfer difficulty.
In summary, our main contributions are:

\vspace{-5pt}
\begin{itemize}[leftmargin=*, itemsep=0pt]
    \item[$\bullet$] We investigate prompt tuning and adapter tuning methods to understand their effectiveness across domains with varying transfer difficulties. 
    Our findings reveal that the efficacy of each method with each modality varies across different of transfer difficulty, with notable performance of VPT and TA for high-difficulty domains.
    \item[$\bullet$] 
    We propose \alg, which utilizes VPT and TA for tuning and employ an adaptive ensemble approach to optimally leverage the general knowledge of VLMs for each domain. The ensemble's coefficient is adaptively determined by the distances to learned classes, serving as an estimate of transfer difficulty.
    \item[$\bullet$] We show that \alg achieves state-of-the-art performance across various downstream tasks, with particularly notable improvements in unseen tasks during adaptation.
\end{itemize}
\vspace{-10pt}
\section{Backgrounds}
Here, we provide a brief overview of the background related to our method. For a detailed explanation with more related works is in Appendix~\ref{appendix:more_related_work}.

\paragraph{\textbf{Zero-shot CLIP.}} CLIP~\cite{radford2021learning} is designed for creating visual features based on natural language guidance.
The CLIP model can perform zero-shot inference, classifying an image into one of $C$ possible classes without additional training. This is achieved by calculating the cosine similarity between an visual feature $\mathbf{z}$, derived from the visual encoder, and the text features of each class $\{ \mathbf{t}_{i}\}_{i=1}^{C}$, which are obtained from the text encoder.

For processing the image, let us define the visual encoder as $\mathcal{V}$, which comprises $L_{\mathcal{V}}$ layers, denoted as $\{\mathcal{V}_{i}\}_{i=1}^{L_{\mathcal{V}}}$. 
The encoder takes patch embeddings $\mathbf{E}_0  \in \mathbb{R}^{M \times d_v}$ as input, which are obtained by dividing the image $I$ into $M$ fixed-size patches. Patch embeddings $\mathbf{E}_i$ is then fed into the $(i+1)^{\text{th}}$ transformer block ($\mathcal{V}_{i+1}$) along with a learnable class ([CLS]) tokens $\mathbf{c}_i$. This process is sequentially carried out through all $L_{\mathcal{V}}$ transformer blocks, formulated as follows:
\begin{align}
    [\mathbf{c}_i, \mathbf{E}_i] &= \mathcal{V}_{i}\left([\textbf{c}_{i-1}, \textbf{E}_{i-1}] \right) \;\; i = 1,\ldots,L_{\mathcal{V}}, \label{eq:img_zeroshot} \\
    \mathbf{z} &= \text{\textbf{\texttt{ImageProj}}}(\textbf{c}_{L_{\mathcal{V}}}) \label{eq:img_proj},
\end{align}
Here, $[\cdot, \cdot]$ denotes the concatenation operation. We can obtain the text features in a similar way with word embeddings $\mathbf{W}_{0} = [\mathbf{w}_{0}^{1}, \ldots, \mathbf{w}_{0}^{N}] \in \mathbb{R}^{N \times d_l}$ and text encoder $\mathcal{T}$ which is consist of $L_{\mathcal{T}}$ layers $\{\mathcal{T}_{i}\}_{i=1}^{L_{\mathcal{T}}}$, as follows:
\begin{align}
    [\textbf{W}_i] &= \mathcal{T}_{i}(\textbf{W}_{i-1}) \;\; i = 1,\ldots, L_{\mathcal{T}} \label{eq:txt_zeroshot} \\
    \mathbf{t}_{i} &= \text{\textbf{\texttt{TextProj}}}(\textbf{w}_{L_{\mathcal{T}}}^{N}) \label{eq:txt_proj}
\end{align}
The predicted probability for class $i$ is as:
\begin{equation}
    \label{eq:predicted_prob}
    \text{Pr}(y=i|\mathbf{z}, \mathbf{t}) = \frac{\exp \left(\text{sim}(\mathbf{z}, \mathbf{t}_{i})/ \tau \right)}{\sum_{j=1}^{C} \exp \left(\text{sim}(\mathbf{z}, \mathbf{t}_{j})/ \tau \right)},
\end{equation}
where $\text{sim}(\cdot, \cdot)$ indicates cosine similarity and $\tau$ is the learned temperature of CLIP. We can also interpret the text features as a \textbf{classifier}~\cite{gao2023clip, zhang2022tip}, where $\mathbf{t}_{i}$ is the classifier weight for class $i$.

\vspace{-5pt}
\paragraph{\textbf{Prompt Tuning for CLIP.}}
To enable prompt tuning~\cite{zhou2022conditional, khattakMaPLe, zhu2023prompt, khattak2023self}, we replace the Eq.~(\ref{eq:img_zeroshot}) and Eq.~(\ref{eq:txt_zeroshot}) by newly introducing $b_{\mathcal{V}}$ and $b_{\mathcal{T}}$ learnable tokens $\{ \hat{P}_{i}^{k} \in \mathbb{R}^{d_v} \}_{k=1}^{b_{\mathcal{V}}}$ and $\{ P_{i}^{k} \in \mathbb{R}^{d_l} \}_{k=1}^{b_{\mathcal{T}}}$ for $i^{\text{th}}$ layer, and their concatenation $\hat{\mathbf{P}}_{i}$ and $\mathbf{P}_{i}$. We can introduce the visual prompt for the first $J_{\mathcal{V}}$ layers of the visual encoder, then we can compute as follows:
\begin{align}
    &[\textbf{c}_i, \textbf{E}_i,\; \text{\underline{\;\;}}] = \mathcal{V}_{i}([\textbf{c}_{i-1}, \textbf{E}_{i-1}, \hat{\textbf{P}}_{i-1}]),
    \label{eq:vpt} \\
    &[\textbf{c}_j, \textbf{E}_j, \hat{\textbf{P}}_{j}] = \mathcal{V}_{j}([\textbf{c}_{j-1}, \textbf{E}_{j-1}, \hat{\textbf{P}}_{j-1}]), \nonumber
\end{align}
for $i = 1, \ldots, J_{\mathcal{V}}$ and $j = J_{\mathcal{V}}+1, \ldots, L_{\mathcal{V}}$. Also, we can replace Eq.~(\ref{eq:txt_zeroshot}) to belows by introducing text prompt for the fisrt $J_{\mathcal{T}}$ layers of text encoder:
\begin{align}\label{eq:lp}
    &[\text{\underline{\;\;}},\!\textbf{W}_{i}] \!=\! \mathcal{T}_{i}([\textbf{P}_{i-1},\!\textbf{W}_{i-1}]) \;\; i \!=\! 1,\!\ldots,\!J_{\mathcal{T}}, \\
    &[\textbf{P}_{j},\!\textbf{W}_{j}] \!=\! \mathcal{T}_{j}([\textbf{P}_{j-1},\!\textbf{W}_{j-1}]) \;\; j \!=\! J_{\mathcal{T}} \!+\! 1,\!\ldots,\!L_{\mathcal{T}}. \nonumber
\end{align}

Here, we train the visual and text prompt for the first $J_{\mathcal{V}}$ and $J_{\mathcal{T}}$ layers of corresponding encoders.

\vspace{-5pt}
\paragraph{\textbf{Adapter-style Tuning for CLIP.}} To enable adapter-style tuning, we replace Eq.~(\ref{eq:img_proj}) and Eq.~(\ref{eq:txt_proj}) by introducing \textbf{\texttt{ImgAdapt}} and \textbf{\texttt{TxtAdapt}} which are shallow stacking networks upon the frozen CLIP model~\cite{gao2023clip, zhang2022tip, zhu2023not}. 
\begin{align}
    & \tilde{\mathbf{z}} = \text{\textbf{\texttt{ImgProj}}}(\mathbf{c}_{L_{\mathcal{V}}}), \;\; \mathbf{z} = \text{\textbf{\texttt{ImgAdapt}}}(\tilde{\mathbf{z}}) 
    \label{eq:va} \\ 
    & \tilde{\mathbf{t}} = \text{\textbf{\texttt{TxtProj}}}(\mathbf{w}_{L_{\mathcal{T}}}^{N}), \;\; \mathbf{t} = \text{\textbf{\texttt{TxtAdapt}}}(\tilde{\mathbf{t}}) \label{eq:la}
\end{align}

\vspace{-10pt}

\definecolor{lp}{RGB}{147,112,219}
\definecolor{ivlp}{RGB}{32,178,170}
\definecolor{vpt}{RGB}{255, 99, 71}

\begin{figure*}[t]
\centering
\small
\begin{minipage}[t]{0.48\textwidth}
    \centering
    \includegraphics[width=1.0\linewidth]{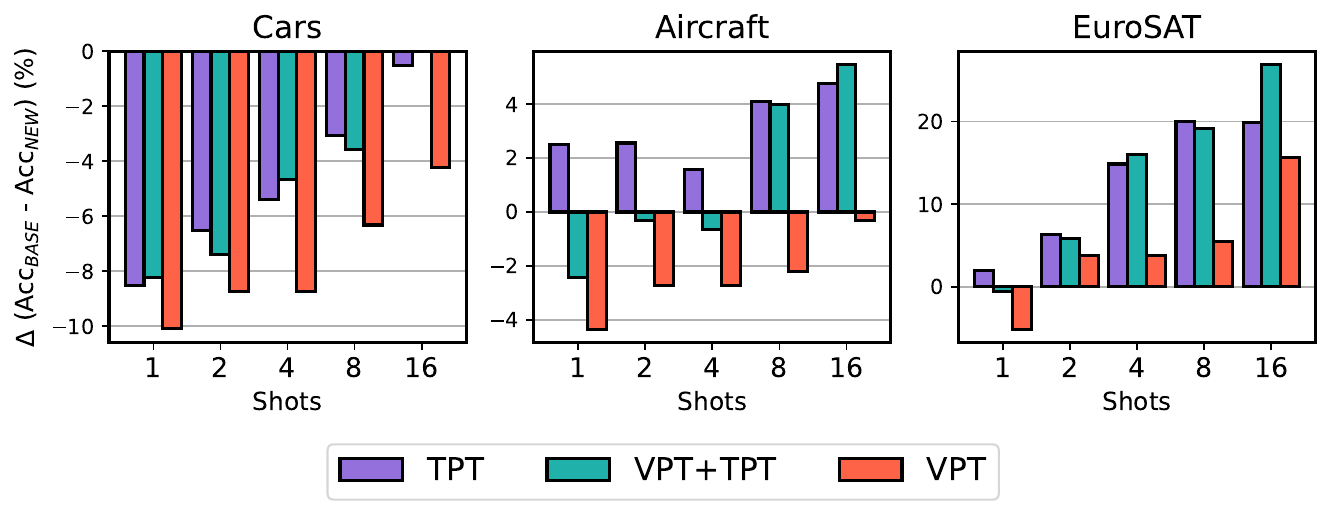}
    \caption{
    Comparison of accuracy differences~(\%) between base and novel categories across three prompt tuning options (\textcolor{lp}{\textbf{TPT}}, \textcolor{ivlp}{\textbf{VPT+TPT}}, \textcolor{vpt}{\textbf{VPT}}) with varying numbers of shots.}
    \label{fig:diff}
\end{minipage}
\hfill
\begin{minipage}[t]{0.48\textwidth}
\centering
\small
\includegraphics[width=1.0\linewidth]{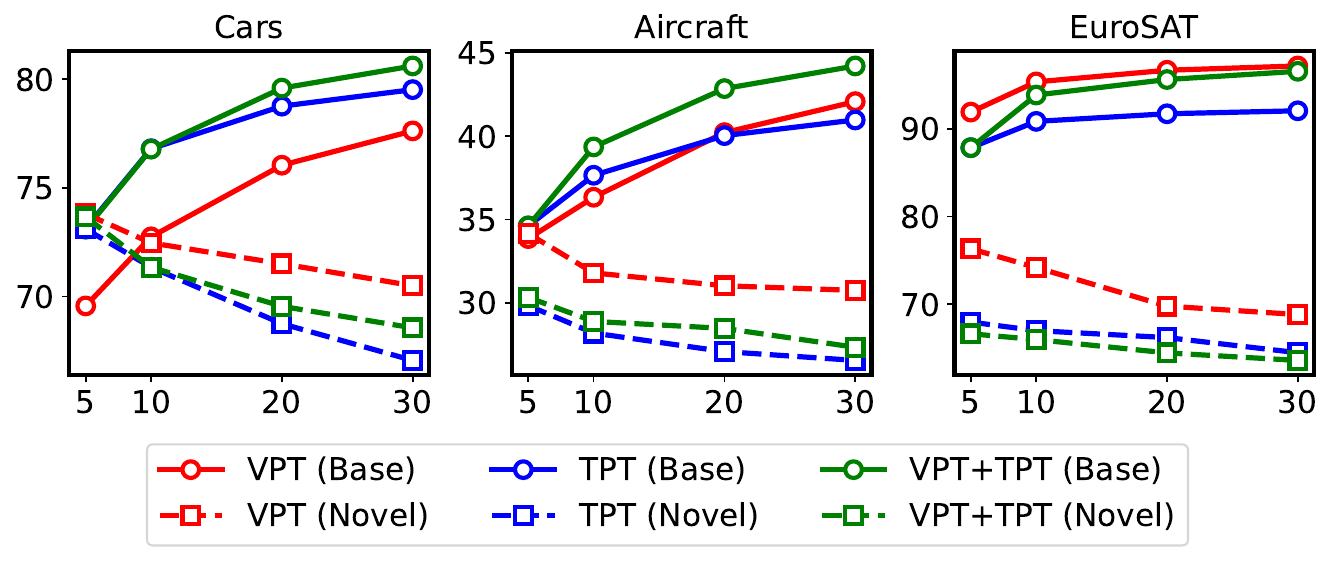}
    \caption{
    Comparison of the accuracy~(\%) of base and novel categories using TPT, VPT, and their combination (VPT+TPT) on three transfer learning datasets over various training epochs. 
    }\label{fig:epoch}
\end{minipage}
\vspace{-5pt}
\end{figure*}
\begin{figure*}[t]
    \centering
    \includegraphics[width=0.9\linewidth]{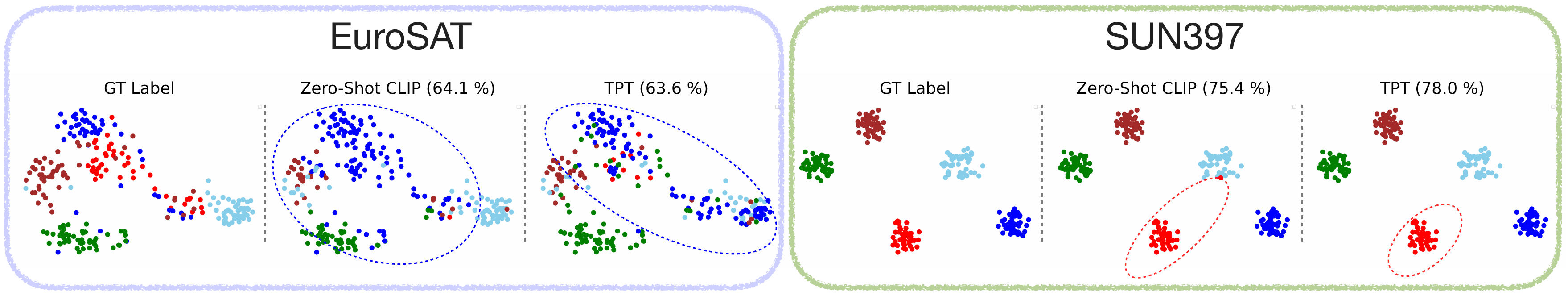}
    \caption{
    t-SNE~\cite{van2008visualizing} plots of visual features for novel category with their corresponding labels~\textcolor{blue}{\textbf{(left)}}, zero-shot CLIP prediction~\textcolor{blue}{\textbf{(middle)}}, and prediction with TPT~\textcolor{blue}{\textbf{(right)}}. 50 samples are randomly selected from each class in EuroSAT and SUN397, using all 5 classes in EuroSAT and 5 randomly chosen classes from SUN397. Dotted lines within the t-SNE plot represent the decision boundaries corresponding to each class, indicated by the same color.
    }
    \label{fig:lp_tsne}
    \vspace{-10pt}
\end{figure*}
\section{Motivating Observations}
\label{sec:observation}
\vspace{-5pt}
Here, we analyze the behavior of visual and text encoders depending on different tuning methods and transfer difficulty of target domains within the framework of ETL.
To accomplish this, we begin by categorizing domains based on their relative transfer difficulty~(RTD), which is a metric first defined by \citet{yu2023task}. 
\begin{definition}[Relative Transfer Difficulty \cite{yu2023task}]
    Let $f(\cdot)$ and $g(\cdot)$ be random classifiers where the precision of each equals $1/C$, and zero-shot CLIP, respectively. Also, $\text{Prec}_{f}$ and $\text{Prec}_{g}$ denote the precision of classifiers $f$ and $g$. Then, RTD is formulated as follows:
    \begin{equation*}
        \text{RTD} = \frac{\text{Prec}_{f}}{\text{Prec}_{g}}=\frac{1/C}{\text{Prec}_{g}} = \frac{1}{C \cdot \text{Prec}_{g}}
    \end{equation*}
\end{definition}

\noindent Under this metric, we identify EuroSAT, DTD, and FGVC Aircraft as the three most challenging domains, while ImageNet, SUN397, and Stanford Cars are recognized as the three easiest domains. We will primarily focus on these six domains to clearly demonstrate the impact of RTD on VLMs' behavior. To assess adaptability and generalizability, we train the CLIP-B/16 utilizing each prompt tuning approach on tasks requiring generalization from base to novel categories. Here, ``base category" refers to a subset of classes within the domain learned through few-shot methods, and ``novel category" is those not included in the training. Each dataset is split into these categories; the model is trained on base classes with 16 shots and tested on both. Therefore, performance on the ``base category" is related to adaptability, and performance in the ``novel category" is related to generalizability. More detailed values are present in Appendix~\ref{appendix:observation_details}.
\begin{observation}\label{obs:obs_1}
     VPT offers better generalizability than TPT. While TPT has greater adaptability to seen classes in low-difficulty domains, it is not effective for high-difficulty domains and shows overfitting to the base classes.
\end{observation}

We commence with an analysis of the separate behavior of visual and text prompts during the tuning process. Fig.~\ref{fig:diff} illustrates the performance discrepancy between the two categories for each method. Across all domains, VPT consistently shows the smallest performance gap for every shot number, indicating reduced overfitting to base classes. This observation is especially prominent in domains with high RTD though the trend is not as pronounced in domains with low RTD. We also observe that combining VPT and TPT does not consistently mitigate the overfitting of TPT, as evidenced by the larger performance gap in FGVC Aircraft and EuroSAT compared to TPT alone.

Fig.~\ref{fig:epoch} displays the comparative performance of base and novel categories over different epochs. 
While all prompt tuning methods show an improvement in base category performance at the expense of generalization, VPT consistently exhibits a lesser decline in novel category performance. 
Notably, for challenging domains like FGVC Aircraft and EuroSAT, VPT exceeds the novel performance of TPT and their combination regardless of epoch. 



\begin{figure*}[t]
\centering
\small
\begin{minipage}[b]{0.49\textwidth}
    \centering
    \includegraphics[width=1.0\linewidth]{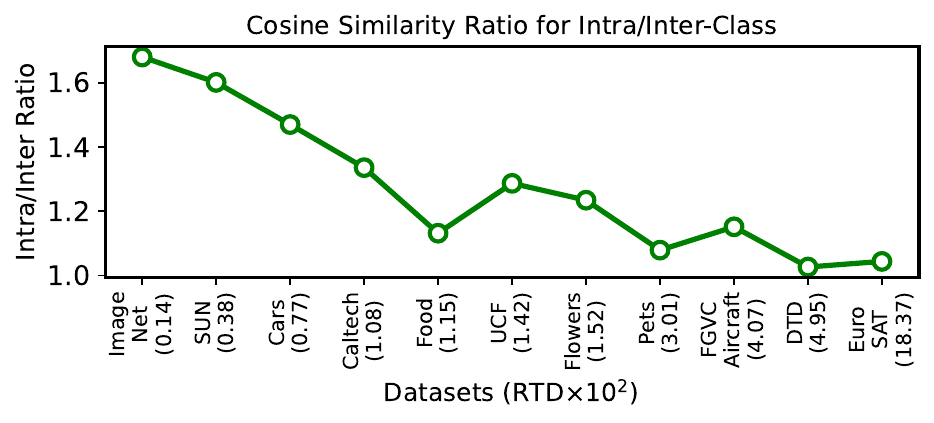}
    \caption{
    Comparison of intra- and inter-class ratios to show class separability across different datasets with their RTD, arranged from low to high RTD.
    }
    \label{fig:intra_inter}
\end{minipage}
\hfill
\begin{minipage}[b]{0.47\textwidth}
\centering
\small
    \includegraphics[width=1.0\linewidth]{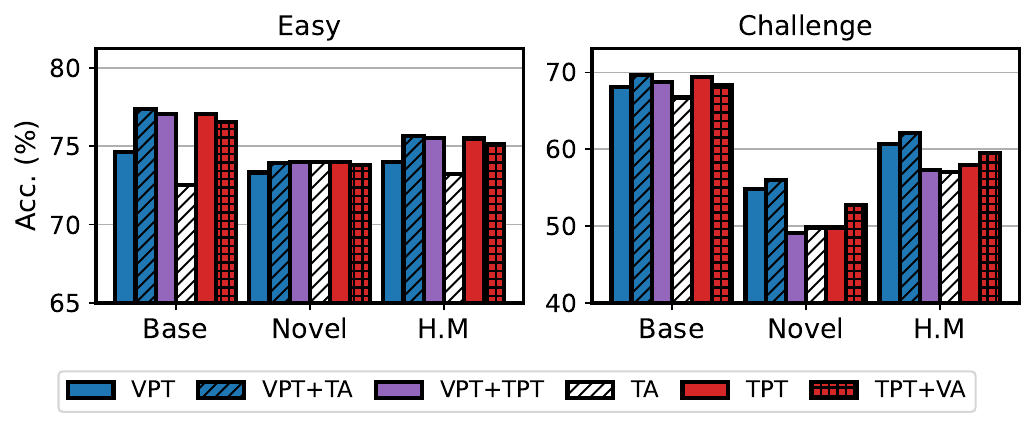}
    \caption{Comparison of the combined effectiveness of prompt tuning and adapter-style tuning. ``Easy" refers to three domains with low RTD, and ``Challenge" refers to three domains with high RTD.
    }
    \label{fig:vp_la}
\end{minipage}
\vspace{-12.5pt}
\end{figure*}
\begin{observation}\label{obs:obs_2}
    Low class separability of visual features is the primary reason for the overfitting of TPT on high RTD.
\end{observation}

Class separability is a critical factor in determining the transferability of a source model to a target domain~\cite{pandy2022transferability}. 
To determine the class separability of visual features, we use the ratio of intra- to inter-class cosine similarities~\cite{oh2021boil, zhu2023not}. Fig.~\ref{fig:intra_inter} demonstrates that the ratio is higher in domains with lower RTD, which are considered easier, and lower in more challenging datasets with higher RTD. These findings suggest that the class separability highly correlates with transfer difficulty, strongly influencing the overfitting risk of TPT on high RTD domains.

To see how class separability affects TPT, we further explore the visual features and predictions of zero-shot CLIP and TPT. As shown in Fig.~\ref{fig:lp_tsne}, EuroSAT, which exhibits a high RTD, shows lower class separability compared to SUN397 that has a lower RTD. 
Furthermore, in EuroSAT, when TPT attempts to classify visual features with low class separability, its performance for novel classes is lower than zero-shot CLIP. This is because TPT tries to fit the decision boundary, represented as dotted lines, to features that are challenging to classify by solely adjusting classifier weights with multiple stacks of learnable prompts. This underscore the significance of separable visual features, a factor closely linked to VPT.
Consequently, this leads to significant overfitting, where the decision boundary of one class overlaps with others.
Conversely, with visual features that exhibit high class separability, TPT's predictions are more accurate than those of zero-shot CLIP as it can easily determine the better decision boundary.%
These results underscore the significance of separable visual features, a factor closely linked to VPT. 

\begin{observation}\label{obs:obs_3}
    TA effectively enhances adaptability with a low risk of overfitting when employed with VPT, especially on higher RTD datasets.
\end{observation}
Fig.~\ref{fig:vp_la} shows that while TA and VPT each exhibit less adaptability than TPT alone, together they outperform across all categories, signifying both high adaptability and generalizability. This advantageous combination is particularly significant for higher RTD, while the performance improvement in novel categories with lower RTD is marginal.
\begin{figure}[t]
    \centering
    \includegraphics[width=1.0\linewidth]{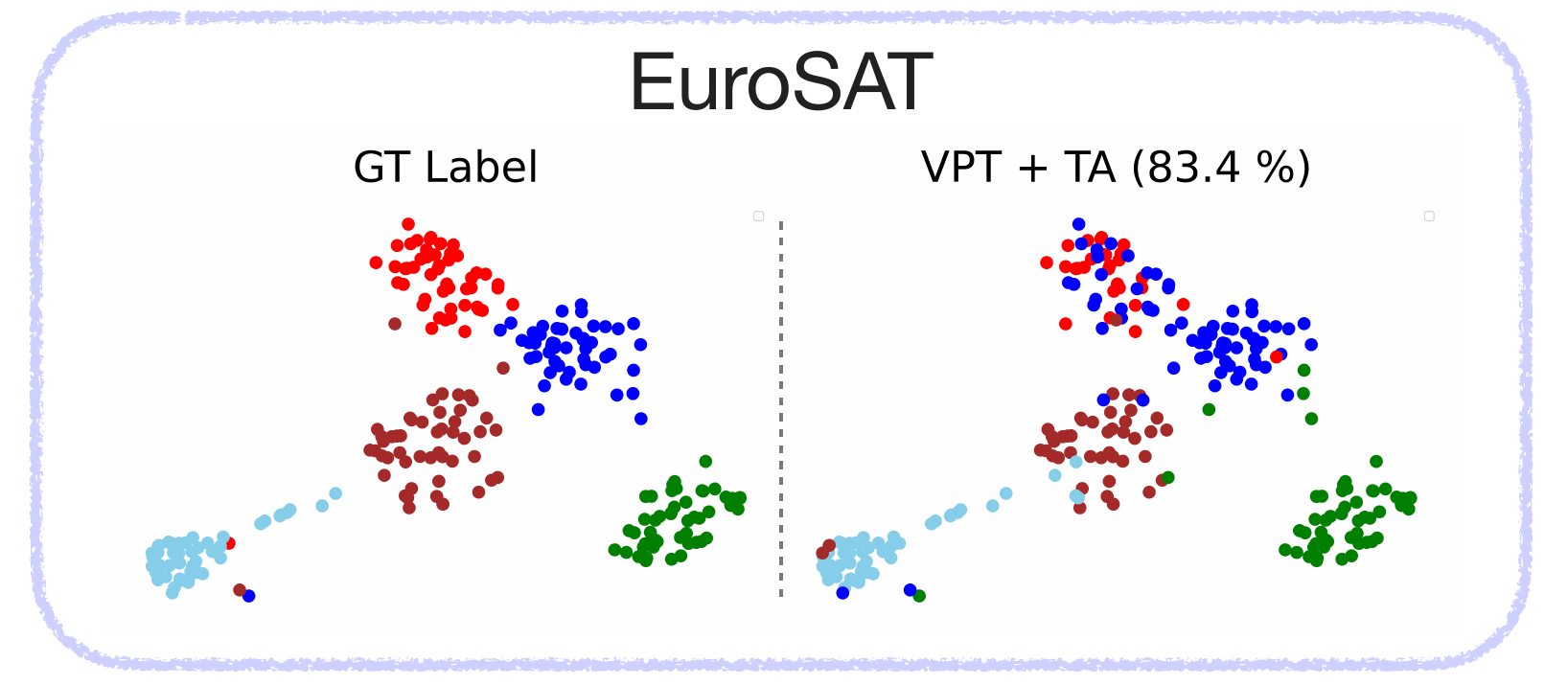}
    \caption{
    t-SNE plots of visual features of CLIP with VPT for a novel category with their corresponding labels \textcolor{blue}{\textbf{(left)}} and prediction with TA \textcolor{blue}{\textbf{(right)}}. 50 samples are randomly selected from each class.
    }
    \label{fig:vptla_tsne}
    \vspace{-10pt}
\end{figure}

This synergy occurs because VPT enhances the class separability in visual features, allowing the linear transformation of classifier weights to suffice for adaptation, as depicted in Fig.~\ref{fig:vptla_tsne}. 
TA simply modifies the features of the pre-trained text encoder, preventing overconfidence in the decision boundary, especially for domains with high RTD and low class separability. 
In addition, we conduct experiments using a combination of TPT and a visual adapter (VA). However, this combination proves less effective than integrating VPT and TA, further emphasizing the importance of visual feature separability.

\begin{observation}\label{obs:obs_4}
    By modulating the influence of TA through an ensemble of pre-adapter and post-adapter features, each with a domain-specific coefficient, we can significantly improve generalization in low RTD domains while maintaining high performance in high RTD domains.
\end{observation}

While combining VPT and TA has great synergy in high RTD domains, utilizing TA can result in the loss of some general knowledge from the original CLIP, which is crucial for domains with low RTD. This is evident in Tab.~\ref{tab:motif_ensemble}, as na\"ively using VPT and TA together may lead to a degradation in performance on novel classes in domains with low RTD. This is because for low RTD, a lot of tasks within the domain need to lie in the region of general knowledge, as illustrated in Fig.~\ref{fig:intro_figure}(b). But the training of a TA creates a task-specific boundary which may not be optimal for other tasks within the same domain. In domains with high RTD, task-specific knowledge gained from adapters can also enhance performance on unseen tasks, as the general knowledge is often insufficient for these domains.

\definecolor{myred}{RGB}{204, 4, 67}
\definecolor{myblue}{RGB}{0,0,205}

\begin{table}[t]
\centering
\caption{
    Comparison of accuracy~(\%) on novel classes between zero-shot CLIP, without an ensemble, an ensemble with fixed coefficient, and an ensemble with optimal coefficient. We determine the fixed coefficient as 0.4, based on average novel performance.
}
\resizebox{\columnwidth}{!}{
\begin{tabular}{l|cccc}
\toprule[0.1em]
Dataset                     & \textbf{SUN397} & \textbf{Stanford Cars} & \textbf{DTD} & \textbf{EuroSAT} \\ \midrule
ZS CLIP                        & 75.35 & 74.89 & 59.90 & 64.05 \\ \midrule
\multirow{2}{*}{VPT + TA}   & 74.52 & 68.40 & 63.05 & 77.73 \\
                            & (\textcolor{myblue}{-0.83}) & (\textcolor{myblue}{-6.49}) & (\textcolor{myred}{+3.15})  & (\textcolor{myred}{+13.68}) \\ \midrule
+ Fixed Ens                 & 78.68 & 74.22 & 64.16 & 75.87 \\ 
($\alpha = 0.4$)            & (\textcolor{myred}{+3.33}) & (\textcolor{myblue}{-0.67}) & (\textcolor{myred}{+4.26})   & (\textcolor{myred}{+11.82}) \\ \midrule
\multirow{2}{*}{+ Opt. Ens} & 78.90 & 75.19 & 64.32 & 77.73 \\ 
                            & (\textcolor{myred}{+3.55}) & (\textcolor{myred}{+0.30}) & (\textcolor{myred}{+4.42})  & (\textcolor{myred}{+13.68}) \\ \midrule
Opt. $\alpha$               & 0.3 & 0.0 & 0.5 & 1.0 \\
\bottomrule[0.1em]
\end{tabular}
}\label{tab:motif_ensemble}
\vspace{-10pt}
\end{table}
This degradation in domains with low RTD can be mitigated by diminishing the influence of TA.
Inspired by the residual connection in adapter-style tuning methods~\cite{zhang2022tip, gao2023clip}, we use an ensemble of pre-adapter and post-adapter features for the text encoder. This ensemble, defined with coefficient $\alpha$, can be expressed as:
\begin{equation}
\label{eq:alpha}
    \mathbf{t} = \alpha \cdot \text{\texttt{\textbf{TxtAdapt}}}(\tilde{\mathbf{t}}) + (1-\alpha) \cdot \tilde{\mathbf{t}}.
\end{equation}

As Tab.~\ref{tab:motif_ensemble} illustrates, the ensemble method improves performance in domains with low RTD. However, using pre-adapter features can yield suboptimal outcomes in more challenging domains. For instance, performance on EuroSAT drops from 77.73\% to 75.87\% when $\alpha$ is set as a fixed coefficient, as domains with high RTD demand more from TA. By optimally setting $\alpha$ for each domain, we consistently outperform zero-shot CLIP across all domains by effectively combining general and task-specific knowledge tailored to each domain's needs.  Observing this optimal coefficient, we note that that more challenging domains typically require a higher coefficient. These findings highlight the necessity of a method to calculate an adaptive coefficient of ensemble, which would modulate TA activation according to domain and its RTD.
\section{Method}\label{sec:method}
\vspace{-5pt}
Based on our observations, we propose a new method, \alg, which is a difficulty-agnostic approach that utilizes an adaptive ensemble with tuning methods including VPT and TA.

\vspace{-5pt}
\subsection{Configuration Design \& Training}
Due to the need for a combination of VPT and TA to achieve adaptability and generalizability in highly difficult domains, we configure the trainable parameters to include multiple stacks of visual prompts, and a linear text adaptation layer following the pre-trained text encoder.
While existing adapter-style methods~\cite{zhang2022tip, zhu2023not, gao2023clip} rely on manually optimized text prompts for different datasets, we use learnable text prompts just for the input because manually creating prompt templates for each domain in the real world is challenging. The learnable text prompts are unnecessary if manual prompts are already well-formed, which is further explained in Section~\ref{sec:exp}.

We extract the visual feature $\mathbf{z}$ using Eq.~(\ref{eq:vpt}) and Eq.~(\ref{eq:img_proj}) and the text feature $\mathbf{t}$ using Eq.~(\ref{eq:lp}) with $J_{\mathcal{T}}=1$ and Eq.~(\ref{eq:la}). 
We apply linear adapter parameterized as matrix $\mathbf{A}$ and bias $\mathbf{b}$ for $\text{\textbf{\texttt{TextAdapter}}}$ in Eq.~(\ref{eq:la}) rather than using bottleneck structure~\cite{zhang2022tip, gao2023clip} based on our results in Fig.~\ref{fig:rank_data}. 
Our adapter can be formulated as follows:
\begin{equation}
\label{eq:linear_adapter}
    \mathbf{t} = \text{\textbf{\texttt{TxtAdapt}}}(\tilde{\mathbf{t}}) \coloneqq \mathbf{A}^{\intercal} \tilde{\mathbf{t}} + \mathbf{b}
\end{equation}

\noindent During the training procedure, our objective is to maximize the predicted probability $\text{Pr}(y=y_{\text{gt}}|\mathbf{z}, \mathbf{t})$ for ground truth label $y_{\text{gt}}$ by using cross-entropy loss $\ell_{\text{CE}}(\mathbf{z}, \mathbf{t}, y_{\text{gt}})$ which is defined as follows:
\begin{equation*}
    \ell_{\text{CE}}(\mathbf{z}, \mathbf{t}, y_{\text{gt}}) = \log \text{Pr}(y=y_{\text{gt}}|\mathbf{z}, \mathbf{t}),
\end{equation*}
where the predicted probability is computed as Eq.~(\ref{eq:predicted_prob}).

\subsection{Adaptive Ensemble for Evaluation}
\begin{figure}[t]
    \centering
    \includegraphics[width=0.9\linewidth]{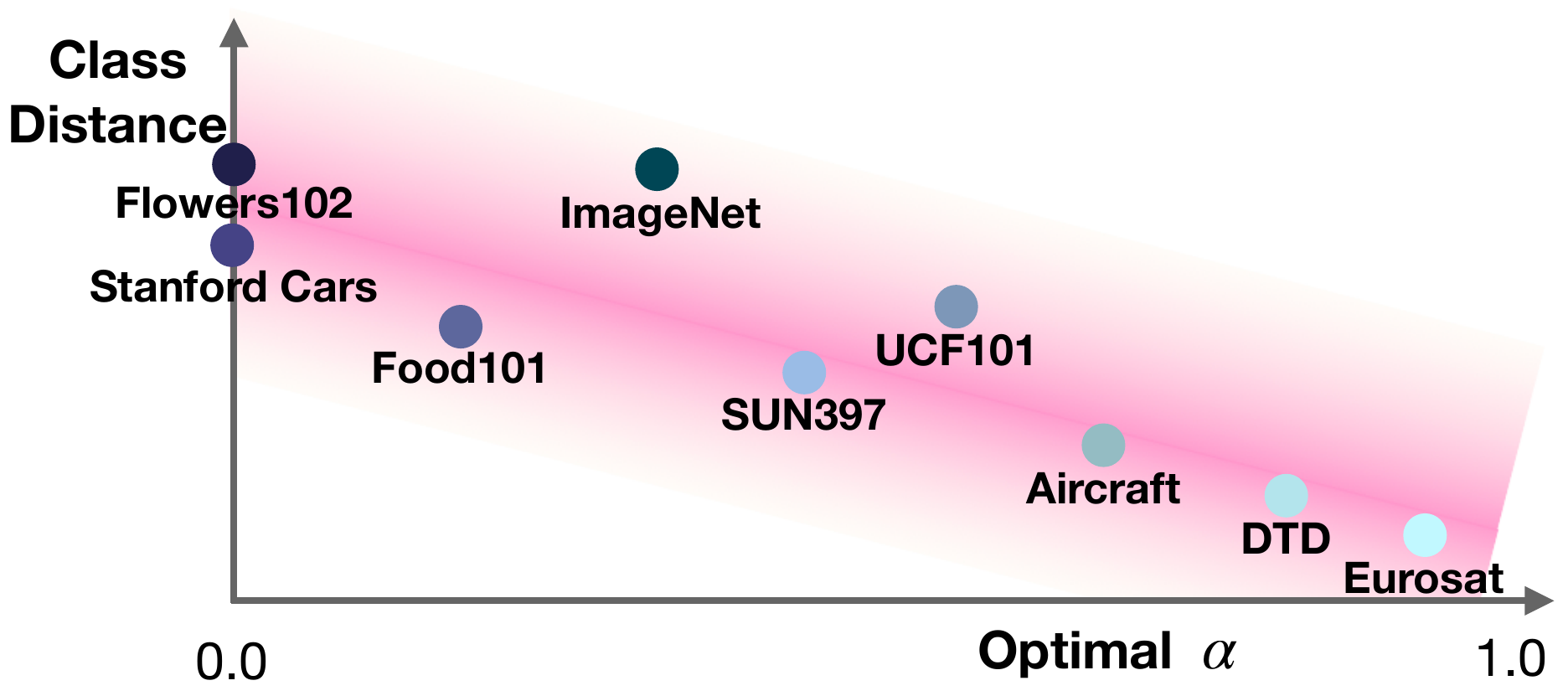}
    \caption{
    The relationship between class distance and optimal $\alpha$ for each domain used in Eq.~\eqref{eq:alpha} and Table~\ref{tab:motif_ensemble}. 
    }
    \label{fig:alpha_graph}
    \vspace{-10pt}
\end{figure}


\begin{figure}[t]
    \centering
    \includegraphics[width=0.9\linewidth]{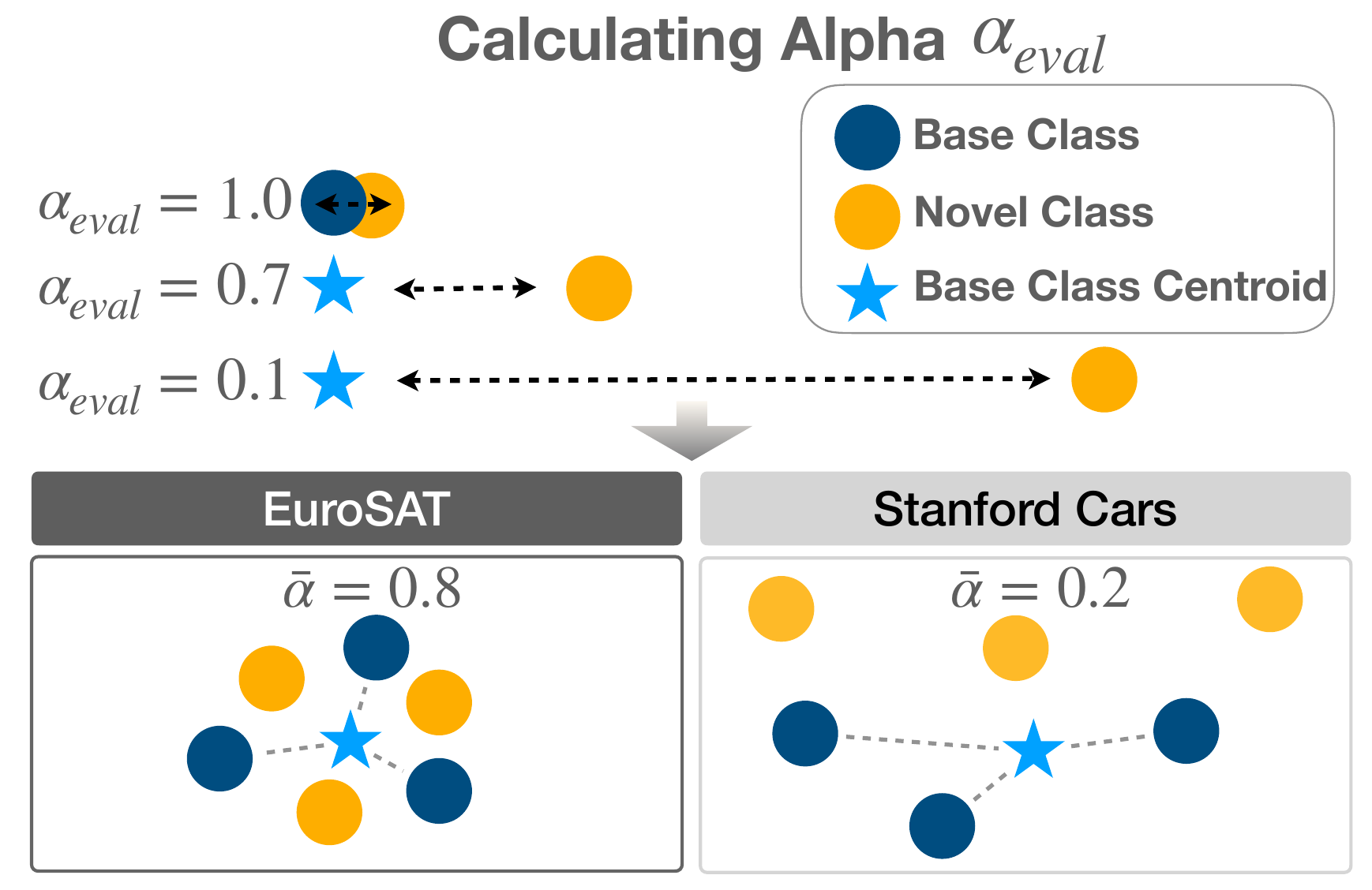}
    \caption{
    A concept figure for calculating the adaptive coefficient $\alpha_{\text{eval}}$ for ensemble upon its class distance.
    }
    \label{fig:eval_figure}
    \vspace{-10pt}
\end{figure}

Due to the various levels of transfer difficulty encountered during deployment, an adaptive method is necessary to avoid suboptimal results for each target domain. Motivated by our observations, in the evaluation stage, we use an adaptive ensemble approach that combines pre-adapter ($\tilde{\mathbf{t}}_{\text{eval}}$) and post-adapter text features (Eq.~\eqref{eq:linear_adapter}), described as follows:
\begin{equation*}
    \mathbf{t}_{\text{eval}} = \alpha_{\text{eval}} \cdot (\mathbf{A}^{\intercal}\tilde{\mathbf{t}}_{\text{eval}} + \mathbf{b}) + (1 - \alpha_{\text{eval}}) \cdot \tilde{\mathbf{t}}_{\text{eval}},
\end{equation*}
\noindent where $\alpha_{\text{eval}}$ is the ensemble coefficient for a target class at evaluation and ${\mathbf{t}}_{\text{eval}}$ is the final representation for that class. With this ensemble approach, for domains with high RTD, the model relies on the adaptability and generalizability of VPT and TA. Conversely, for domains with low RTD, it leverages general knowledge from the pre-trained model to avoid excessive adaptation.
 
To determine the optimal $\alpha_{\text{eval}}$ for each class, which estimates transfer difficulty and acts as a controller for adaptation, we employ a non-parametric method based on the distance between the text features of the evaluation class and the classes learned during training. 
This approach is based on the assumption that in domains with high RTD, class features are typically less separable in the text embedding space, similarly to their separability in the image embedding space. Hence, domains like EuroSAT exhibit low class distances, while those with low RTD, such as Stanford Cars, display high class distances. Fig.~\ref{fig:alpha_graph} shows that the optimal $\alpha$, used in Eq.~\eqref{eq:alpha} and Tab.~\ref{tab:motif_ensemble}, is highly correlated with the distance between class features. 
This tendency suggests that $\alpha_{\text{eval}}$ based on the distance between class features can effectively represent transfer difficulty.

Moreover, instead of applying a single $\alpha_{\text{eval}}$ for all classes, we adopt a class-wise approach. This is because, within the same domain, target features considered as out-of-task should rely more on the general knowledge of pre-trained VLMs, whereas features closer to the learned classes should leverage more task-specific knowledge. 
With regard to this, we adaptively set $\alpha_{\text{eval}}$ by comparing the text feature of the evaluation class with the features of the learned classes, as illustrated in Fig.~\ref{fig:eval_figure}. 
Specifically, we calculate both the average and nearest distances between the evaluation class and the $C$ learned classes in the following manner:
\begin{align*}
    \textstyle d^{\text{avg}}_{\text{eval}} = 1.0 - \frac{1}{C} \sum_{j=1}^{C} \text{sim}({\mathbf{t}'}_{\text{eval}}, {\mathbf{t}'}_{j}), \\
    d^{\text{nn}}_{\text{eval}} = 1.0 - \min_{\forall j \in \{1, \ldots, C\}} \text{sim}({\mathbf{t}'}_{\text{eval}}, {\mathbf{t}'}_{j}),
\end{align*}
\noindent where ${\mathbf{t}'}_{\text{eval}}$ and ${\mathbf{t}'}_{j}$ indicate text feature of evaluation class and learned class $j \in \{1, \ldots, C\}$ from pre-trained VLMs and $\text{sim}$ denotes cosine similarity. Using these distance metrics, we compute the coefficient $\alpha_{\text{eval}}$ as follows:
\vspace{-2.5pt}
\begin{equation*}
    \alpha_{\text{eval}} = \text{exp}\left(-\beta\cdot (d^{\text{avg}}_{\text{eval}}) \cdot \mathbf{1}_{(d^{\text{nn}}_{\text{eval}} > \epsilon)}\right),
\end{equation*}
\vspace{-2.5pt}
\begin{table}[tp]
\centering
\tiny
\caption{Accuracy comparison on base-to-novel generalization of \alg with previous methods. 
}
\vspace{-5pt}
\resizebox{1.0\columnwidth}{!}{
\setlength{\tabcolsep}{3.7pt}{
{
\begin{tabular}{p{1.5cm}l|cccccc}
\toprule[0.1em]
\textbf{Dataset}
& & \textbf{CLIP} & {\begin{tabular}[c]{@{}c@{}}\textbf{CLIP}\\[-0.4ex]
\textbf{-Adapter} \end{tabular}} & {\begin{tabular}[c]{@{}c@{}}\textbf{Co}\\[-0.4ex]
\textbf{-CoOp} \end{tabular}} & \textbf{MaPLe} & {\begin{tabular}[c]{@{}c@{}}\textbf{Pro}\\[-0.4ex]
\textbf{-Grad} \end{tabular}}  & \textbf{APEX} \\
\midrule
\multirow{3}{1.5cm}{Average on 11 datasets} & Base & 69.34 & 83.23 & 81.11 & 82.52 & 82.55 & \textbf{83.99} \\
& Novel & 74.22 & 70.13 & 70.55 & 74.24 & 72.20 & \textbf{76.76} \\
& HM & 71.70 & 75.64 & 75.03 & 77.86 & 76.77 & \textbf{80.04} \\

\midrule
\multirow{3}{1.5cm}{ImageNet} & Base & 72.43 & 76.06 & 76.47 & 77.02 & 76.97 & \textbf{77.12} \\
& Novel & 68.14 & 68.40 & 69.60 & 70.15 & 67.20 & \textbf{71.10} \\
& HM & 70.22 & 72.03 & 72.87 & 73.42 & 71.75 & \textbf{73.99} \\

\midrule
\multirow{3}{1.5cm}{Caltech101} & Base & 96.84 & 98.00 & 97.70 & 97.95 & 97.88 & \textbf{98.18} \\
& Novel & 94.00 & 93.66 & 93.96 & 94.60 & 93.57 & \textbf{95.06} \\
& HM & 95.40 & 95.78 & 95.78 & 96.25 & 95.68 & \textbf{96.59} \\

\midrule
\multirow{3}{1.5cm}{OxfordPets} & Base & 91.17 & 94.86 & 95.66 & \textbf{95.80} & 95.00 & {95.11} \\
& Novel & 97.26 & 94.49 & 96.32 & \textbf{97.82} & 97.46 & 97.27 \\
& HM & 94.12 & 94.67 & 95.99 & \textbf{96.80} & 96.21 & {96.18} \\

\midrule
\multirow{3}{1.5cm}{Stanford Cars} & Base & 63.37 & 77.62 & 72.92 & 74.69 & 78.64 & \textbf{80.53} \\
& Novel & 74.89 & 68.53 & 71.98 & 73.53 & 70.23 & \textbf{75.08} \\
& HM & 68.65 & 72.79 & 72.45 & 74.11 & 74.20 & \textbf{77.71} \\

\midrule
\multirow{3}{1.5cm}{Flowers102} & Base & 72.08 & 96.88 & 94.82 & 95.90 & 94.83 & \textbf{97.47} \\
& Novel & \textbf{77.80} & 69.20 & 70.71 & 72.96 & 74.70 & {77.58} \\
& HM & 74.83 & 80.73 & 81.01 & 82.87 & 83.57 & \textbf{86.40} \\

\midrule
\multirow{3}{1.5cm}{Food101} & Base & 90.10 & 90.02 & \textbf{90.63} & {90.46} & 90.40 & 89.60 \\
& Novel & 91.22 & 89.76 & 91.13 & 91.71 & 90.43 & \textbf{92.06} \\
& HM & 74.83 & 89.89 &  90.88 & \textbf{91.08} & 90.41 & {90.81} \\

\midrule
\multirow{3}{1.5cm}{FGVC Aircraft} & Base & 27.19 & 40.14 & 36.19 & 37.76 & 40.77 & \textbf{42.69} \\
& Novel & \textbf{36.29} &31.77 & 26.82 & 34.67 & 30.16 & {35.21} \\
& HM & 31.09 & 35.47 & 30.81 & 36.15 & 34.67 & \textbf{38.59} \\

\midrule
\multirow{3}{1.5cm}{SUN397} & Base & 69.36 & \textbf{81.72} & 80.55 & 81.33 & 81.19 & 81.17 \\
& Novel & 75.35 & 73.54 & 75.48 & 77.75 & 73.42 & \textbf{78.98} \\
& HM & 72.23 & 77.41 & 77.93 & 79.50 & 77.11 & \textbf{80.06} \\

\midrule
\multirow{3}{1.5cm}{DTD} & Base & 53.24 & 81.77 & 77.34 & 79.34 & 76.64 & \textbf{82.45} \\
& Novel & 59.90 & 49.02 & 48.86 & 56.64 & 54.23 & \textbf{63.80} \\
& HM & 56.37 & 61.29 & 59.89 & 66.10 & 63.52 & \textbf{71.94} \\

\midrule
\multirow{3}{1.5cm}{EuroSAT} & Base & 56.48 & 91.55 & {87.05} &  \textbf{93.00} & 91.23 & 92.83 \\
& Novel & 64.05 & 61.10 & 61.27 & 69.17 & 68.58 & \textbf{79.89} \\
& HM & 60.03 & 73.29 & 71.92 & 79.33 & 78.30 & \textbf{85.88} \\

\midrule
\multirow{3}{1.5cm}{UCF101} & Base & 70.53 & \textbf{86.87} & 82.86 & 84.43 & 84.54 & {86.74} \\
& Novel & 77.50 & 71.94 & 69.92 & 77.64 & 74.24 & \textbf{78.37} \\
& HM & 73.85 & 78.70 & 75.84 & 80.89 & 79.06 & \textbf{82.34} \\

\bottomrule[0.1em]
\end{tabular}}}}
\label{table:base_to_new}
\vspace{-10pt}
\end{table}
\noindent where $\beta$ is a scaling factor. The equation indicates a preference for pre-adapter features when the text feature distance from learned classes is large, and for trained TA when it is small. The condition of $d^{\text{nn}}_{\text{eval}} > \epsilon$, where $\epsilon$ is a small value set at $0.05$, serves to treat an evaluation class that is very similar to the base class as identical. This adaptive $\alpha_{\text{eval}}$ enables flexible use of general and task-specific knowledge. Moreover, since text embeddings are usually pre-calculated~\cite{radford2021learning}, this adaptive coefficient incurs only a minor computational overhead.

\vspace{-5pt}
\paragraph{\textbf{Vision Ensemble.}} Additionally, to further improve the performance by leveraging more general knowledge of the pretrained VLMs, we can also employ an ensemble technique for the visual encoder that combines the visual feature of the pre-trained VLM ($\mathbf{z}'$) with the task-adapted VLMs ($\mathbf{z}$) as follows:
\vspace{-10pt}
\begin{equation*}
    \mathbf{{z}} = \bar{\alpha} \cdot \mathbf{z'} + (1 - \bar{\alpha}) \cdot \mathbf{z},
\end{equation*}
$\bar{\alpha}$, the mean value of $\alpha_{\text{eval}}$, is used for image ensemble since class-specific $\alpha_{\text{eval}}$ cannot be applied at the image level. 


\vspace{-5pt}
\section{Experiments}\label{sec:exp}
We describe our experimental setup and results for verifying superiority of our method. Additional experimental results are described in Appendix~\ref{appendix:additional_experiments}.

\vspace{-5pt}
\subsection{Experimental Setup}

\paragraph{Datasets.}
We evaluate \alg on the three most commonly used transfer learning tasks: base-to-novel generalization, cross-dataset evaluation, and domain generalization.
For all the few-shot experiments except domain generalization, we follow CoCoOp \cite{zhou2022conditional} which uses 11 image recognition datasets. 
The datasets cover multiple recognition tasks including ImageNet~\cite{inproceedings} and Caltech101~\cite{FeiFei2004LearningGV} which consists of generic objects; OxfordPets~\cite{Parkhi2012CatsAD}, Stanford Cars~\cite{Krause20133DOR}, Flowers102~\cite{Nilsback2008AutomatedFC}, Food101~\cite{Bossard2014Food101M}, and FGVC Aircraft~\cite{Maji2013FineGrainedVC} for fine-grained classification, SUN397~\cite{Xiao2010SUNDL} for scene recognition, UCF101~\cite{Soomro2012UCF101AD} for action recognition, DTD~\cite{Cimpoi2013DescribingTI} for texture classification, and EuroSAT~\cite{Helber2017EuroSATAN} which consists of satellite images. 
For the domain generalization benchmark, we use ImageNet as a source dataset and use ImageNet-A~\cite{Hendrycks2019NaturalAE}, ImageNet-R~\cite{Hendrycks2020TheMF}, ImageNet-Sketch~\cite{Wang2019LearningRG}, and ImageNetV2~\cite{Recht2019DoIC} as out-of-domain datasets.

\paragraph{Experimental Details.}
We use multiple baselines for comparison with our methods in experiments. 
These include the standard zero-shot CLIP~\cite{radford2021learning}, CLIP-Adapter~\cite{gao2023clip}, CoCoOp~\cite{zhou2022conditional} and MaPLe~\cite{khattakMaPLe}. We also consider ProGrad~\cite{zhu2023prompt}, which uses gradient alignment for prompt learning. 
When reporting results, we have reproduced all the experiments, as we observe that the values are highly dependent on the random seed. Instead of taking the average results from three seeds, as done in previous works \cite{khattakMaPLe}, we use the \textbf{average of 20 seeds} to determine the final value for base-to-novel and the \textbf{average of 5 seeds} for cross-evaluation and domain-generalization. Additionally, we found that using the Adadelta optimizer \cite{zeiler2012adadelta} yields better results, so we have reproduced the experiments with Adadelta. 
More experimental details can be found in the Appendix~\ref{appendix:implementation_details}.


\begin{table}[t]
\centering
\caption{Comparison of accuracy on cross-dataset of \alg with previous methods.}
\vspace{-5pt}
\resizebox{\columnwidth}{!}{
\begin{tabular}{l|l|cccccc}
\toprule[0.15em]
\multicolumn{2}{c|}{\textbf{Dataset}} & \textbf{C-Adapter} & \textbf{CoCoOp} &  \textbf{MaPLe} & \textbf{ProGrad} & \alg \\
\midrule
\textbf{Source} & ImageNet & 70.12 & 71.46 & 70.58 & 71.73 &  \textbf{72.00} \\ \midrule\multirow{10}{*}{\textbf{Target}} & Caltech101 & 92.94 & 93.24 & 93.46 & 93.30 & \textbf{94.46} \\
 & OxfordPets & 86.80 & \textbf{90.38} & 90.28 & 89.95 & 90.06 \\
 & Cars & 64.22 & 64.08 & 65.22 & 65.25 & \textbf{65.46} \\
 & Flower102 & 69.06 & 70.50 & \textbf{71.80} & 69.34 & 71.58 \\
 & Food101 & 85.20 & 85.64 & 86.24 & 86.22 & \textbf{86.44} \\
 & Aircraft & 24.24 & 21.58 & 23.62 & 21.22 & \textbf{24.44} \\
 & SUN397 & 64.36 & 66.30 & \textbf{67.32} & 65.32 & 67.20 \\
 & DTD & 43.44 & 43.68 & 45.04 & 42.19 & \textbf{45.70} \\
 & EuroSAT & \textbf{47.66} & 45.48 & 46.24 & 45.33 & 47.58 \\
 & UCF101 & 65.52 & 67.42 & 68.26 & 67.62 & \textbf{68.80} \\
 \midrule
 \multicolumn{2}{c|}{\textbf{Average}} & 64.34 & 64.83 & 65.75 & 64.57 & \textbf{66.16} \\
\bottomrule[0.15em]
\end{tabular}}
\label{table:cross_eval}
\end{table}

\begin{table}[t]
\centering
\small
\caption{Comparison of accuracy on domain generalization of \alg with previous methods.}
\vspace{-5pt}
\setlength{\tabcolsep}{2.5pt}{
\begin{tabular}{lc*{5}{c}}
\toprule[0.12em]
    \multirow{2}{*} & {\textbf{Source}} & \multicolumn{5}{c}{\textbf{Target}} \\
    \cmidrule(l{2pt}r{2pt}){2-2} \cmidrule(l{2pt}r{2pt}){3-7} 
    & {ImageNet} & {-V2} & {-S} & {-A} & {-R} & {Avg.}  \\  \midrule
    C-Adapter & 70.12 & 61.78 & 46.70 & 48.56 & 74.00 & 57.76 \\
    CoCoOp & 71.46 & 64.44 & 48.58 & 50.20 & 75.64 & 59.72 \\
    MaPLe & 70.58 & 63.95 &  \textbf{48.78} & 50.53 & \textbf{76.78} & 59.90 \\
    ProGrad & 71.73 & 64.54 & 48.59 & 50.38 & 75.87 & 59.85 \\ \midrule
    \alg & \textbf{72.00} & \textbf{64.70} & 48.48 &  \textbf{50.68} & 76.76 &  \textbf{60.16} \\
\bottomrule[0.12em]
\end{tabular}}
\label{table:generalization}
\vspace{-10pt}
\end{table}
\subsection{Main Results}
\paragraph{Base-to-Novel Generalization.} In this scenario, the datasets are evenly divided into base and novel categories. The model is trained on the base classes using 16 shots and is subsequently tested on both the base and novel classes. 
As indicated in Table~\ref{table:base_to_new}, \alg consistently outperforms the best of the previous methods in average accuracy across all datasets, with a margin of 1$\sim$6\%. In particular, our method exhibits superior performance in novel classes on all datasets, demonstrating \alg's enhanced generalizability. The exceptions are Oxford Pets and FGVC Aircraft, where the performance is already exceptionally high and low, respectively. This improvement is especially notable in domains with high RTD, such as EuroSAT ($+15.84\%$) and DTD ($+3.90\%$). Additionally, the \alg method also shows superior performance in base categories, highlighting the high adaptability of our approach.

\vspace{-5pt}
\paragraph{Cross-dataset Evaluation.}
We train the model to generalize across different domains by using a cross-dataset evaluation task. Specifically, we first train the model on the ImageNet dataset and then transfer it to the 10 other datasets. Table~\ref{table:cross_eval} summarizes that \alg shows the best overall performance compared to existing baselines. Our proposed method achieves the best performance on 7 out of 11 tasks. This demonstrates \alg's effectiveness, especially in difficult situations where both the task and domain are unseen.

\vspace{-5pt}
\paragraph{Domain Generalization.}
We assess the capability of \alg to generalize to out-of-distribution data by training on the source dataset, ImageNet, and subsequently testing on various modified versions of ImageNet. Our method does not achieve a large margin of superiority since our adaptive ensemble is primarily designed to enhance performance in novel classes. Nonetheless, our method still surpasses all baseline models on average accuracy in this domain generalization task.

\subsection{Ablation Study}
In this section, we provide ablation experiments on \alg. Full results are detailed in Appendix~\ref{appendix:additional_experiments}.

\vspace{-5pt}

\begin{table}[t]
\centering
\caption{
    Comparison of the effect of adaptive ensemble technique between text and visual encoder by RTD.
}
\resizebox{0.9\columnwidth}{!}{
\begin{tabular}{cc|ccc}
\toprule[0.1em]
Text & Visual & Easy & Challenge & All \\ \midrule
\xmark & \xmark & 70.67 & 58.25 & 74.61 \\
\cmark & \xmark & 74.51 (\textcolor{myred}{+3.84}) & 58.66 (\textcolor{myred}{+0.41}) & 76.19 (\textcolor{myred}{+1.58}) \\
\xmark & \cmark & 70.79 (\textcolor{myred}{+0.12}) & 58.65 (\textcolor{myred}{+0.40}) & 74.83 (\textcolor{myred}{+0.22}) \\
\cmark & \cmark & \textbf{75.05} (\textcolor{myred}{+4.38}) & \textbf{59.63} (\textcolor{myred}{+1.38}) & \textbf{76.76} (\textcolor{myred}{+2.15}) \\
\bottomrule[0.1em]
\end{tabular}
}\label{tab:ensemble}
\vspace{-5pt}
\end{table}

\paragraph{Effect of Ensemble.} We have conducted a component analysis of two adaptive ensemble techniques of \alg, focusing on (1) the text encoder and (2) the visual encoder. The results, as shown in Table~\ref{tab:ensemble}, reveal that the ensembling of the text encoder is crucial for enhancing performance. Conversely, ensembling the visual encoder results in a minor yet consistent improvement. The text ensemble notably achieves substantial improvements in domains with low RTD, implying that task-specific knowledge is primarily acquired through TA. Overall, employing both ensemble techniques leads to the most improvement regardless of RTD.

\vspace{-5pt}
\begin{figure}[t]
    \centering
    \includegraphics[width=1.0\linewidth]{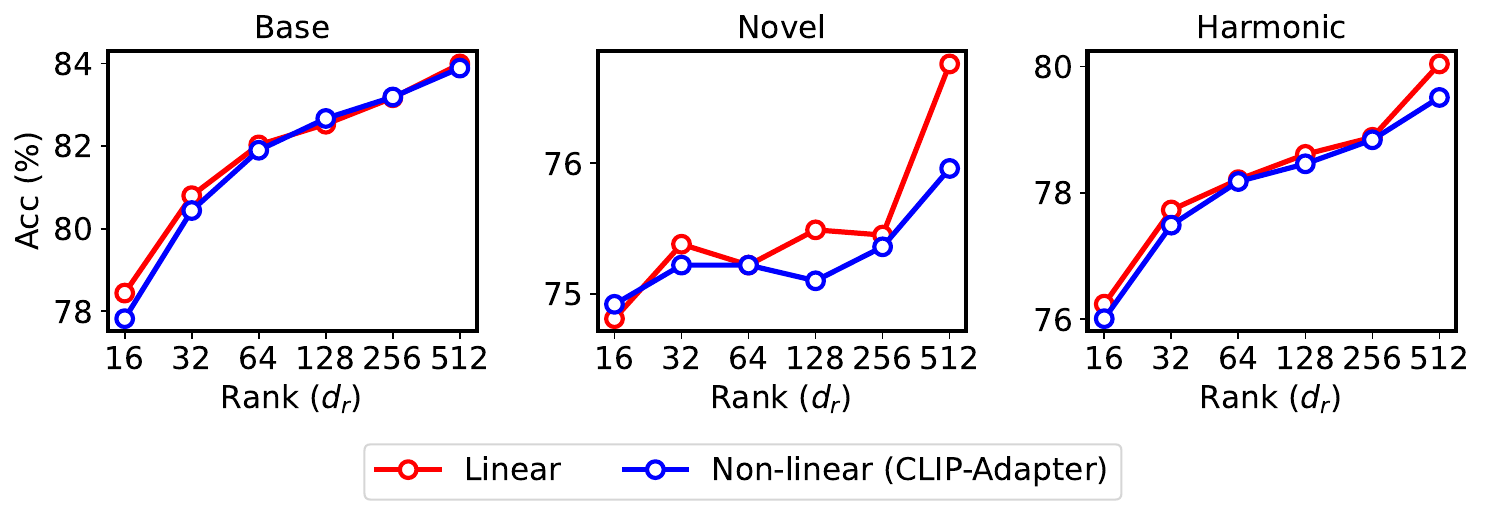}
    \caption{
    Comparison of the accuracy of base, novel, and their harmonic mean using low-rank linear adapter and bottleneck layer of non-linear adapter~\cite{gao2023clip}.
    }
    \label{fig:rank}
\vspace{-10pt}
\end{figure}

\paragraph{Using Low-Rank Linear Adapter.} CLIP-Adapter~\cite{gao2023clip} and Tip-Adapter~\cite{zhang2022tip} utilize the bottleneck layer~\cite{he2016deep} which shrinks and re-expands the feature dimensions to improve efficiency. Similarly, we utilize low-rank matrix factorization that $\mathbf{A} = \mathbf{U}\mathbf{V}^{\intercal}$ where $\mathbf{V}, \mathbf{U} \in \mathbb{R}^{d_{l} \times d_{r}}$ with $d_{r} < d_{l}$ to improve the parameter efficiency. Fig.~\ref{fig:rank} shows that although TA's performance diminishes with decreasing dimension $d_{r}$, average accuracy with few parameters ($d_{r} = 32$) still achieves performance comparable to ProGrad~(\citealt{zhu2023prompt}; +0.72\%). Moreover, the linear adapter consistently outperforms the non-linear adapter~\cite{gao2023clip} across all values of $d_{r}$, motivating us to use a linear adapter in our proposed \alg.


\section{Conclusion}
We propose \alg to address the challenges of conventional prompt and adapter-style ETL methods for VLMs. Our approach incorporates two key components based on our observations: (1) using VPT and TA for exploiting the property of each modality and (2) adaptive ensemble coefficient in the inference stage. We empirically demonstrate the superior performance of \alg, consistently achieving a better performance than the previous methods.

\section*{Acknowledgements}
This work was supported by Institute of Information \& communications Technology Planning \& Evaluation (IITP) grant funded by the Korea government (MSIT) (No.2019-0-00075, Artificial Intelligence Graduate School Program (KAIST), 5\%), Institute of Information \& communications Technology Planning \& Evaluation (IITP) grant funded by the Korea government(MSIT) (No. RS-2024-00457882, AI Research Hub Project), 5\%), and Institute of Information \& communications Technology Planning \& Evaluation (IITP) grant funded by the Korea government(MSIT) (No.2022-0-00641, XVoice: Multi-Modal Voice Meta Learning, 90\%).

\section*{Limitation}
We focus on two types of ETL, prompt tuning and adapter-style tuning, for VLMs for vision-language understanding tasks such as CLIP, EVA-CLIP, and CoCA-CLIP. While our extensive analyses provide valuable insights, our paper primarily centers on understanding tasks, with opportunities for further exploration in vision-language generation tasks such as BLIP~\cite{li2022blip} and LLaVA~\cite{liu2024visual}. Additionally, though we focus on two main representative ETL methods, further analyses could be conducted on other ETL methods like LoRA~\cite{hu2022lora} and IA3~\cite{liu2022fewshot}. We leave these aspects for future work but wish to emphasize the comprehensive exploration provided by our study on the two representative ETL methods for VLMs.



\bibliographystyle{acl_natbib}
\bibliography{main}

\clearpage
\appendix
\section{Implementation Details}
\label{appendix:implementation_details}


As explained in Section~\ref{sec:exp}, we utilize the ViT-B/16 model as the CLIP image encoder and a standard GPT2-like structure with an End Of Text (EOT) token as the classification token for the text encoder. To implement \alg, we use visual prompts for all layers, setting $J_{V} = 12$ for base-to-novel generalization and $J_{V} = 3$ for cross-evaluation and domain generalization. The text prompt is applied only to the shallow prompt, and therefore, $J_{V} = 1$ for all experiments. The number of prompts for each layer, $b_{\mathcal{V}}$ and $b_{\mathcal{T}}$, is set to 2. The initial text prompt is fixed as \textit{``a photo of a"}, and the visual prompts are initialized with a zero-mean Gaussian distribution with a standard deviation of $0.02$. The matrix term of the text adapter is initialized with an identity matrix, and the bias vector is initialized with a zero vector.


For training, we use the Adadelta optimizer~\cite{zeiler2012adadelta} with a learning rate of $0.15$ and a cosine learning rate scheduler. The batch size is set to $16$, and we train for $15$ epochs, except for ImageNet, where we train for 5 epochs. As in previous works, we apply augmentation techniques of random cropping and flipping. The scaling factor $\beta$, used for calculating $\alpha_{eval}$, is set to $4.0$. In the SGD experiments presented in Appendix~\ref{appendix:additional_experiments}, we adopt a batch size of $16$ and epochs of $30$ and $5$ for ImageNet, along with a learning rate of $0.0015$ and a cosine learning rate scheduler. The augmentation and scaling factors are set the same as in the Adadelta experiments.

For reproducing baselines, we use the Adadelta optimizer with a learning rate of 0.25, selected after a grid search with values $[0.1,0.15,0.2,0.25,0.3]$. The rest of the settings remain the same as in the original papers. Results with their original configurations using SGD optimizer are listed in Appendix~\ref{appendix:additional_experiments}. All our experiments were conducted on a single NVIDIA RTX 3090. 


\newpage
\section{Notation and Algorithm}
\label{appendix:notation_algorithm}

In this section, we present the notation and algorithm of our method, \alg. The notation is detailed in Table~\ref{table:notation}. The training algorithm for \alg is outlined in Algorithm~\ref{alg:overview}, and the adaptive inference algorithm is presented in Algorithm~\ref{alg:inference}.
\begin{table}[h]
\centering
\tiny
\caption{The notation table for Section~\ref{sec:observation}
}
\resizebox{0.9\columnwidth}{!}{
\begin{tabular}{c|l}
\toprule[0.1em]
\textbf{Notation} & \multicolumn{1}{c}{\textbf{Description}} \\ \midrule
\multicolumn{2}{l}{\textit{The notation for VLMs}} \\ \midrule
$\mathcal{V}$ & The visual encoder of VLMs \\
$\mathcal{T}$ & The text encoder of VLMs \\
$L_{\mathcal{V}}$ & The number of layers of visual encoder \\
$L_{\mathcal{T}}$ & The number of layers of text encoder \\ 
$\mathcal{V}_{\ell}$ & The $\ell$$^{\text{th}}$ Transformer layer of visual encoder \\
$\mathcal{T}_{\ell}$ & The $\ell$$^{\text{th}}$ Transformer layer of text encoder \\
$\mathbf{E}_{\ell}$ & The patch embeddings of $\ell$$^{\text{th}}$ layer of visual encoder \\
$\mathbf{W}_{\ell}$ & The word embeddings of $\ell$$^{\text{th}}$ layer of text encoder \\
\midrule
\multicolumn{2}{l}{\textit{The inputs for VLMs or prompt tuning}} \\ \midrule
$J_{\mathcal{V}}$ & The number of layers of VPT \\
$J_{\mathcal{T}}$ & The number of layers of TPT \\ 
$b_{\mathcal{V}}$ & The context length of VPT \\
$b_{\mathcal{T}}$ & The context length of TPT \\ 
$\hat{\mathbf{P}}_{\ell}$ & The visual prompt of $\ell$$^{\text{th}}$ layer of visual encoder \\
$\mathbf{P}_{\ell}$ & The text prompt of $\ell$$^{\text{th}}$ layer of text encoder \\
\midrule
\multicolumn{2}{l}{\textit{The outputs for VLMs}} \\ \midrule
$\mathbf{c}_{\ell}$ & The embedded features of $\ell$$^{\text{th}}$ layer for [CLS] token \\
$\mathbf{t}_{i}$ & The text features of $i$$^{\text{th}}$ class \\
$\mathbf{z}$ & The visual features from visual encoder \\
\midrule
\multicolumn{2}{l}{\textit{The outputs for VLMs related to \alg}} \\ \midrule
\multirow{2}{*}{$\mathbf{z}'$} & The visual features from visual encoder of \\
& pretrained VLMs for adaptive ensemble \\
\multirow{2}{*}{$\mathbf{t}'$} & The text features from text encoder of \\
& pretrained VLMs for adaptive ensemble \\
\multirow{2}{*}{$\tilde{\mathbf{t}}$} & The pre-adapter text features of text encoder of \\
& adapted VLMs \\
\bottomrule[0.1em]
\end{tabular}}
\label{table:notation}
\vspace{-10pt}
\end{table}


\begin{algorithm}[ht]
\caption{Pseudo-Algorithm for Training of \alg}
\label{alg:overview}
\begin{algorithmic}[1]
\REQUIRE Pretrained visual encoder $\mathcal{V}$, Pretrained text encoder $\mathcal{T}$, Learnable vision prompts $\hat{\mathbf{P}}$, Shallow text prompts $\mathbf{P_0}$, Adapter parameterized by matrix $\mathbf{A}$ and $\mathbf{b}$ \\
\REQUIRE Training Samples $\mathcal{S}$, Initial Text Embeddings $\textbf{W}_{0}$ \\
\STATE Randomly initialize $\phi = [\hat{\mathbf{P}}, \mathbf{A}, \mathbf{b}]$\\
 \WHILE{not done}
    \STATE Sample Batch $\mathcal{B} = (I, y_{gt})$
    \STATE $\mathbf{E}_0 = \text{\texttt{PathEmbedding(I)}}$
    \FOR{$i = 1, \ldots, J_{\mathcal{V}}$}
        \STATE $[\mathbf{c}_i, \mathbf{E}_i, \text{\underline{\;\;\;\;}}] \gets \mathcal{V}_{i}([\mathbf{c}_{i-1}, \mathbf{E}_{i-1}, \hat{\mathbf{P}}_{i-1}])$
    \ENDFOR
    \FOR{$i = J_{\mathcal{V}} + 1, \ldots, L_{\mathcal{V}}$}
        \STATE $[\mathbf{c}_i, \mathbf{E}_i, \hat{\mathbf{P}}_i] \gets \mathcal{V}_{i}([\mathbf{c}_{i-1}, \mathbf{E}_{i-1}, \hat{\mathbf{P}}_{i-1}])$
    \ENDFOR
    \STATE $\mathbf{z} \gets \text{\texttt{ImageProj}}(\mathbf{c}_{L_{\mathcal{V}}})$
    \STATE $\mathbf{\tilde{t}} = \mathcal{T}([\mathbf{W_0}, \mathbf{P_0}])$
    \STATE $\mathbf{t} = \mathbf{A}^{\intercal}\mathbf{\tilde{t}} + \mathbf{b}$
    \STATE \textcolor{darkgray}{/* Calculate the probability for class $i$ */}
    \STATE $\text{Pr}(y=i|\mathbf{z}, \mathbf{t}) = \frac{\exp(\text{sim}(\mathbf{z}, \mathbf{t}_{i})/ \tau)}{\sum_{j=1}^{C} \exp(\text{sim}(\mathbf{z}, \mathbf{t}_{j})/ \tau)}$
    \STATE $\ell_{\text{CE}}(\mathbf{z}, \mathbf{t}, y_{\text{gt}}) = \log \text{Pr}(y=y_{\text{gt}}|\mathbf{z}, \mathbf{t})$
    \STATE $\phi = \phi - \gamma\nabla_{\phi}{\ell}_{\text{CE}}(\mathbf{z}, \mathbf{t}, y_{\text{gt}} ; \phi)$
\ENDWHILE







\end{algorithmic}
\end{algorithm}

\begin{algorithm}[ht]
\caption{Pseudo-Algorithm for Adaptive Inference of \alg}
\label{alg:inference}
\begin{algorithmic}[1]
\REQUIRE Pretrained visual encoder $\mathcal{V}$, Pretrained text encoder $\mathcal{T}$, Learned vision prompts $\hat{\mathbf{P}}$, Learned shallow text prompts $\mathbf{P_0}, $Learned adapter parameterized by matrix $\mathbf{A}$ and $\mathbf{b}$, The $C$ classes for base category $\{ 1, \ldots, C \}$, The $C_{\text{eval}}$ candidate classes for evaluation $\{C+1, \ldots, C + C_{\text{eval}}\}$,  \\
\REQUIRE Initial Trained Text Embeddings $\{\mathbf{W}_{0,j}\}_{j=1}^{C}$, Initial Evaluation Text Embedding $\left\{\textbf{W}_{0,\text{eval}} \right\}_{\text{eval}=C+1}^{C+C_{\text{eval}}}$, Evaluation Image $I$ \\
\STATE $\{\mathbf{{t}'}_{j}\}_{j=1}^{C} = \{\mathcal{T}(\mathbf{W}_{0,j})\}_{j=1}^{C}$

\FOR{$\text{eval} = C+1, \ldots, C+C_{\text{eval}}$}

\STATE $\mathbf{{t}'}_{\text{eval}} = \mathcal{T}(\mathbf{W}_{0,\text{eval}})$

\STATE $\mathbf{\tilde{t}}_\text{eval} = \mathcal{T}([\textbf{W}_{0,\text{eval}}, \mathbf{P_0}])$

\STATE $d^{\text{avg}}_{\text{eval}} = 1.0 - \frac{1}{C} \sum_{j=1}^{C} \text{sim}({\mathbf{t}'}_{\text{eval}}, {\mathbf{t}'}_{j})$
\STATE $d^{\text{nn}}_{\text{eval}} = 1.0 - \min_{\forall j \in \{1, \ldots, C\}} \text{sim}({\mathbf{t}'}_{\text{eval}}, {\mathbf{t}'}_{j})$

\STATE $\alpha_{\text{eval}} = \text{exp}\left(-\beta\cdot (d^{\text{avg}}_{\text{eval}}) \cdot \mathbf{1}_{(d^{\text{nn}}_{\text{eval}} > \epsilon)}\right)$

\STATE $\mathbf{t}_{\text{eval}} = \alpha_{\text{eval}} \cdot ( \mathbf{A}^{\intercal} \tilde{\mathbf{t}}_{\text{eval}} + \mathbf{b}) + (1-\alpha_{\text{eval}}) \cdot \tilde{\mathbf{t}}_{\text{eval}}$

\ENDFOR

\STATE $\mathbf{E}_0 = \text{\texttt{PathEmbedding}}(I)$
\STATE ${\mathbf{c}'}_{L_{\mathcal{V}}} = \mathcal{V}([\mathbf{c}_0, \mathbf{E_0}])$
\STATE $\mathbf{z}' \gets \text{\texttt{ImageProj}}({\mathbf{c}'}_{L_{\mathcal{V}}})$

\FOR{$i = 1, \ldots, J_{\mathcal{V}}$}
    \STATE $[\mathbf{c}_i, \mathbf{E}_i, \text{\underline{\;\;\;\;}}] \gets \mathcal{V}_{i}([\mathbf{c}_{i-1}, \mathbf{E}_{i-1}, \hat{\mathbf{P}}_{i-1}])$
\ENDFOR
\FOR{$i = J_{\mathcal{V}} + 1, \ldots, L_{\mathcal{V}}$}
    \STATE $[\mathbf{c}_i, \mathbf{E}_i, \hat{\mathbf{P}}_i] \gets \mathcal{V}_{i}([\mathbf{c}_{i-1}, \mathbf{E}_{i-1}, \hat{\mathbf{P}}_{i-1}])$
\ENDFOR
\STATE $\mathbf{z} \gets \text{\texttt{ImageProj}}(\mathbf{c}_{L_{\mathcal{V}}})$

\STATE $\bar{\alpha} = \frac{1}{C_{\text{eval}}} \sum_{\text{eval}=C+1}^{C+C_{\text{eval}}} \alpha_{\text{eval}}$

\STATE $\mathbf{{z}} = \bar{\alpha} \cdot \mathbf{z'} + (1 - \bar{\alpha}) \cdot \mathbf{z}$

\STATE \textcolor{darkgray}{/* Calculate the probability for class $i$ */}
\STATE Calculate $\text{Pr}(y=i|\mathbf{z}, \mathbf{t}) = \frac{\exp(\text{sim}(\mathbf{z}, \mathbf{t}_{i})/ \tau)}{\sum_{j=C+1}^{C+C_{\text{eval}}} \exp(\text{sim}(\mathbf{z}, \mathbf{t}_{j})/ \tau)}$

\STATE Predict as $\arg\max_{i \in \{C+1, \ldots, C+C_{\text{eval}}\}} \text{Pr}(y=i|\mathbf{z}, \mathbf{t})$

\end{algorithmic}
\end{algorithm}

\newpage
\section{Additional Experiments}
\label{appendix:additional_experiments}

\subsection{Ablation on Adaptive Ensemble}
\vspace{-5pt}
\begin{table}[ht]
\centering
\small
\caption{Comparison of the effect of adaptive ensemble technique between text and visual encoder by RTD.}
\resizebox{0.9\columnwidth}{!}{
\begin{tabular}{l|cccc}
\toprule[0.1em]
 \textbf{Visual} & \xmark & \xmark & \cmark & \cmark (\alg) \\
\textbf{Text} & \xmark & \cmark & \xmark & \cmark (\alg) \\
\midrule
{ImageNet} & 69.08 & 70.09 &  69.22 &  \textbf{71.10} \\ 
 {Caltech101} & 94.91 & 94.80  & 95.01  & \textbf{95.06} \\
  {OxfordPets} & 97.24 &  \textbf{97.39}  & 97.07 & 97.27 \\
  {Cars} & 68.40 & 74.46 & 68.32 & \textbf{75.08} \\
 {Flower102} & 73.71 & 76.40 & 74.43 &  \textbf{77.58} \\
 {Food101} & 90.70 & 91.83  & 90.82 &  \textbf{92.06} \\
  {Aircraft} &  33.97 &33.89  & 33.87 & \textbf{35.21} \\
 {SUN397} & 74.52 & \textbf{78.98}  & 74.82 & \textbf{78.98} \\
 {DTD} & 63.05 & 63.05 & \textbf{63.82}  & 63.80 \\
 {EuroSAT} & 77.73 & {79.04}  & 78.25 & \textbf{79.89} \\
 {UCF101} & 77.39 & 78.17 & 77.55 &  \textbf{78.37} \\
\midrule
{\textbf{Average}} & 74.61 & 76.19 & 74.83 & \textbf{76.76}  \\
\bottomrule[0.1em]
\end{tabular}}
\label{table:full_ensemble}
\vspace{-5pt}
\end{table}

Table~\ref{table:full_ensemble} illustrates the complete results of the component analysis of the adaptive ensemble. We only display results for novel classes, as these ensemble components do not affect the results for base classes, given that $\alpha_{eval}$ is set to $1.0$ for seen classes. AThe ensemble of the text encoder is crucial as its removal leads to a significant performance drop in domains with low RTD, such as Stanford Cars and SUN397. This demonstrates that moderating TA with an adaptive ensemble helps to leverage both task-specific knowledge and general VLMs knowledge effectively. The ensemble on the visual encoder offers marginal improvement, but combining both still yields the most superior performance on average. 

\subsection{Results on Low-Rank Experiments}
\begin{figure*}[t]


    \centering
    \includegraphics[width=1.0\linewidth]{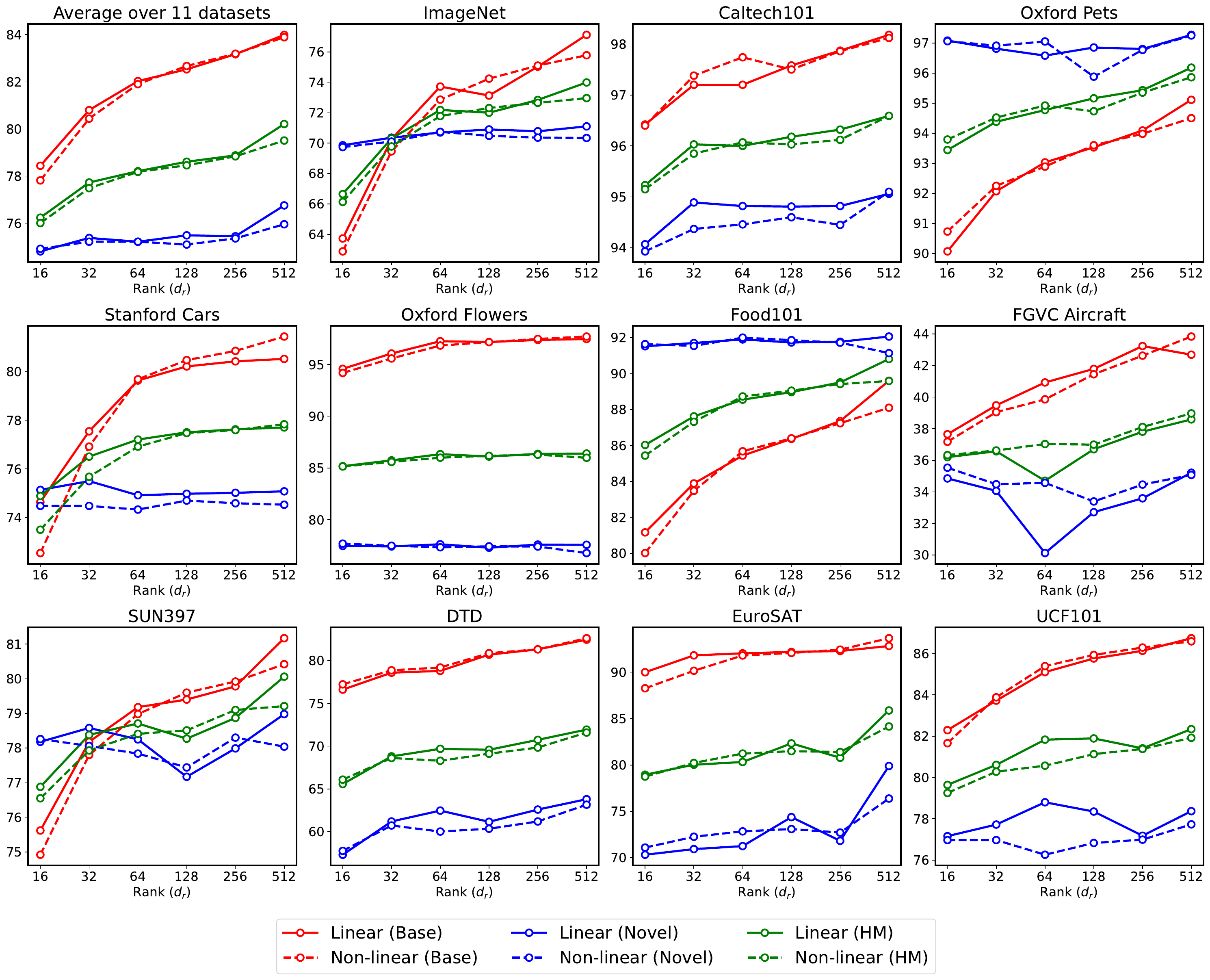}
    \caption{
    Results for the performance of the low-rank approach with different ranks.
    }
    \label{fig:rank_data}
\end{figure*}
\begin{table*}[t]
\centering
\caption{Full results on each dataset of Table~\ref{tab:shallow_prompt}}
\resizebox{\textwidth}{!}{
\setlength{\tabcolsep}{1.8pt}{
\begin{tabular}{ll|cccccccccccc}
\toprule[0.1em]
& & \begin{tabular}[c]{@{}c@{}}\textbf{Average on}\\[-0.5ex]
\textbf{11 datasets} \end{tabular} & \textbf{ImageNet} & \textbf{Caltech101}  & \textbf{OxfordPets} &  \begin{tabular}[c]{@{}c@{}}\textbf{Stanford}\\[-0.5ex]
\textbf{Cars} \end{tabular}  & \textbf{Flowers102} & \textbf{Food101} & \begin{tabular}[c]{@{}c@{}}\textbf{FGVC}\\[-0.5ex]
\textbf{Aircraft} \end{tabular} & \textbf{SUN397} & \textbf{DTD} & \textbf{EuroSAT}  & \textbf{UCF101} \\
\midrule
\multirow{3}{*}{Opt. manual prompt~\cite{zhang2022tip}} & Base & \textbf{84.15} & 76.64 & 98.15 & 95.05 & \textbf{80.75} & 97.45 & 89.35 & \textbf{42.92} & 81.24 & \textbf{83.02} & \textbf{93.93} & 87.10 \\
& Novel & 75.24 & 69.00 & 94.33 & 97.04 & 75.32 & \textbf{77.66} & 91.28 & \textbf{36.42} & 77.60 & 57.59 & 71.74 & \textbf{79.70} \\
& HM & 79.17 & 72.62 & 96.20 & 96.03 & 77.94 & \textbf{86.44} & 90.30 & \textbf{39.40} & 79.38 & 68.01 & 81.35 & \textbf{83.24} \\
\midrule

\multirow{3}{*}{Ens. (60 manual prompts)~\cite{radford2021learning}} & Base & 84.02 & 76.48 & 98.15 & 95.09 & 80.70 & 97.37 & 89.56 & 42.56 & \textbf{81.46} & 82.62 & 93.01 & \textbf{87.18} \\
& Novel & 76.17 & 70.24 & 93.93 & 96.44 & \textbf{75.88} & 77.16 & 91.20 & 35.64 & 78.36 & 59.45 & \textbf{80.35} & 79.21 \\
& HM & 79.70 & 73.23 & 95.99 & 95.76 & \textbf{78.22} & 86.09 & 90.37 & 38.79 & 79.88 & 69.15 & \textbf{86.22} & 83.00 \\
\midrule

\multirow{3}{*}{Shallow prompt~(\alg)} & Base & 83.99 & \textbf{77.12} & \textbf{98.18} & \textbf{95.11} & 80.53 & \textbf{97.47} & \textbf{89.60} & 42.69 & 81.17 & 82.45 & 92.83 & 86.74 \\
& Novel & \textbf{76.76} & \textbf{71.10} & \textbf{95.06} & \textbf{97.27} & 75.08 & 77.58 & \textbf{92.06} & 35.21 & \textbf{78.98} & \textbf{63.80} & 79.89 & 78.37  \\
& HM & \textbf{80.04} & \textbf{73.99} & \textbf{96.59} & \textbf{96.18} & 77.71 & 86.40 & \textbf{90.81} & 38.59 & \textbf{80.06} & \textbf{71.94} & 85.88 & 82.34 \\
\bottomrule[0.1em]
\end{tabular}}}
\label{table:full_shallow_prompt}
\end{table*}

Figure~\ref{fig:rank_data} presents detailed results for each dataset using low-rank methods. The result demonstrates that our linear adapter provides better overall results, particularly for novel classes across most datasets. This parameter-efficient approach exhibits relative robustness in performance, even outperforming MaPLe~\cite{khattakMaPLe} for rank $64$~(+0.32\%) on average. These encouraging results have led us to adopt the linear adapter for the text encoder. Furthermore, we observe that initializing the adapter with an identity matrix improves performance, a strategy that can be explored more thoroughly in future work.

\subsection{Full Results on Manual Text Prompts}
\begin{table}[t]
\centering
\tiny
\caption{ Comparison of baselines using their own configuration (SGD optimizer) with our method.
}
\resizebox{1.0\columnwidth}{!}{
\setlength{\tabcolsep}{1.8pt}{
\begin{tabular}{p{1.5cm}l|cccccc}
\toprule[0.1em]
\textbf{Dataset}
& & \textbf{CLIP} & {\begin{tabular}[c]{@{}c@{}}\textbf{CLIP}\\[-1ex]
\textbf{-Adapter} \end{tabular}} & {\begin{tabular}[c]{@{}c@{}}\textbf{Co}\\[-1ex]
\textbf{-CoOp} \end{tabular}} & \textbf{MaPLe} & {\begin{tabular}[c]{@{}c@{}}\textbf{Pro}\\[-1ex]
\textbf{-Grad} \end{tabular}}  & \textbf{\alg} \\
\midrule
\multirow{3}{1.5cm}{Average on 11 datasets} & Base & 69.34 & 81.81 & 80.28 & 81.74 & 81.78 & \textbf{84.04} \\
& Novel & 74.22 & 71.43 & 72.03 & 73.89 & 69.42 & \textbf{75.67} \\
& HM & 71.70 & 75.93 & 75.60 & 77.30 & 74.80 & \textbf{79.42} \\

\midrule
\multirow{3}{1.5cm}{ImageNet} & Base & 72.43 & 74.40 & 75.99 & 76.81 & \textbf{76.93} & \textbf{76.93} \\
& Novel & 68.14 & 68.63 & 70.39 & \textbf{70.66} & 69.51 & 69.61 \\
& HM & 70.22 & 71.40 & 73.08 & \textbf{73.61} & 73.03 & 73.09 \\

\midrule
\multirow{3}{1.5cm}{Caltech101} & Base & 96.84 & 97.61 & 97.64 & 95.61 & 95.41 & \textbf{98.18} \\
& Novel & 94.00 & 93.72 & 94.52 & 94.71 & 94.05 & \textbf{95.02} \\
& HM & 95.40 & 95.63 & 96.05 & 96.18 & 95.90 & \textbf{96.57} \\

\midrule
\multirow{3}{1.5cm}{OxfordPets} & Base & 91.17 & 95.06 & 95.56 & \textbf{95.61} & 95.41 & 95.21 \\
& Novel & 97.26 & 95.02 & 97.52 & 97.63 & 90.56 & \textbf{97.74} \\
& HM & 94.12 & 95.04 & 96.53 & \textbf{96.61} & 92.92 & 96.46 \\

\midrule
\multirow{3}{1.5cm}{Stanford Cars} & Base & 63.37 & 76.18 & 70.97 & 72.49 & 77.41 & \textbf{80.44} \\
& Novel & \textbf{74.89} & 69.30 & 73.44 & 73.46 & 70.92 & 74.76 \\
& HM & 68.65 & 72.58 & 72.18 & 72.97 & 74.02 & \textbf{77.50} \\

\midrule
\multirow{3}{1.5cm}{Flowers102} & Base & 72.08 & 96.27 & 93.88 & 95.49 & 95.34 & \textbf{97.73} \\
& Novel & \textbf{77.80} & 69.92 & 72.56 & 72.55 & 76.84 & 76.67 \\
& HM & 74.83 & 81.01 & 81.85 & 82.45 & 85.10 & \textbf{85.93} \\

\midrule
\multirow{3}{1.5cm}{Food101} & Base & 90.10 & 90.32 & \textbf{90.54} & 90.50 & 90.17 & 89.46 \\
& Novel & 91.22 & 90.10 & 91.15 & 91.71 & 85.53 & \textbf{91.94} \\
& HM & 74.83 & 90.21 & 90.84 & \textbf{91.10} & 87.79 & 90.68 \\

\midrule
\multirow{3}{1.5cm}{FGVC Aircraft} & Base & 27.19 & 38.87 & 33.64 & 36.33 & 39.01 & \textbf{42.96} \\
& Novel & \textbf{36.29} & 31.95 & 26.49 & 32.64 & 27.77 & 34.72 \\
& HM & 31.09 & 35.07 & 29.64 & 34.39 & 32.44 & \textbf{38.40} \\

\midrule
\multirow{3}{1.5cm}{SUN397} & Base & 69.36 & 76.50 & 79.86 & 80.65 & \textbf{81.35} & 81.18 \\
& Novel & 75.35 & 74.60 & 76.51 & \textbf{78.33} & 69.06 & 77.08 \\
& HM & 72.23 & 75.54 & 78.15 & \textbf{79.47} & 74.70 & 79.08 \\

\midrule
\multirow{3}{1.5cm}{DTD} & Base & 53.24 & 80.46 & 76.58 & 79.20 & 77.45 & \textbf{82.19} \\
& Novel & 59.90 & 52.79 & 53.47 & 55.01 & 51.63 & \textbf{61.21} \\
& HM & 56.37 & 63.75 & 62.97 & 64.92 & 61.96 & \textbf{70.17} \\

\midrule
\multirow{3}{1.5cm}{EuroSAT} & Base & 56.48 & 88.48 & 86.18 & 90.38 & 84.88 & \textbf{93.48} \\
& Novel & 64.05 & 67.12 & 63.04 & 68.43 & 56.66 & \textbf{75.88} \\
& HM & 60.03 & 76.33 & 72.82 & 77.89 & 67.96 & \textbf{83.77} \\

\midrule
\multirow{3}{1.5cm}{UCF101} & Base & 70.53 & 85.81 & 82.22 & 84.02 & 83.82 & \textbf{86.71} \\
& Novel & 77.50 & 72.55 & 73.22 & 77.62 & 71.13 & \textbf{77.77} \\
& HM & 73.85 & 78.62 & 77.46 & 80.69 & 76.96 & \textbf{82.00} \\

\bottomrule[0.1em]
\end{tabular}}}
\label{table:full_sgd}
\vspace{-10pt}
\end{table}
\begin{table}[t]
\centering
\tiny
\caption{Extended baselines not presented in Table~\ref{table:base_to_new} for comparison between base-to-novel experiments with our method. 
}
\resizebox{1.0\columnwidth}{!}{
\setlength{\tabcolsep}{1.8pt}{
\begin{tabular}{p{1.5cm}l|cccccc}
\toprule[0.1em]
\textbf{Dataset}
& & \textbf{CLIP} & \textbf{VPT} & \textbf{TPT} & {\begin{tabular}[c]{@{}c@{}}\textbf{VPT}\\[-1ex]
\textbf{+ TPT} \end{tabular}} & {\begin{tabular}[c]{@{}c@{}}\textbf{Prompt}\\[-1ex]
\textbf{-SRC} \end{tabular}}  & \textbf{APEX} \\
\midrule
\multirow{3}{1.5cm}{Average on 11 datasets} & Base & 69.34 & 81.01 & 82.07 & 82.93 & \textbf{84.36} & {83.99} \\
& Novel & 74.22 & 73.11 & 73.90 & 74.15 & 75.37 & \textbf{76.76} \\
& HM & 71.70 & 76.55 & 77.51 & 78.00 & 79.39 & \textbf{80.04} \\

\midrule
\multirow{3}{1.5cm}{ImageNet} & Base & 72.43 & 75.94 & 76.81 & 77.18 & \textbf{77.90} & {77.12} \\
& Novel & 68.14 & 68.74 & 69.45 & 69.86 & 70.26 & \textbf{71.10} \\
& HM & 70.22 & 72.16 & 72.94 & 73.34 & 73.88 & \textbf{73.99} \\

\midrule
\multirow{3}{1.5cm}{Caltech101} & Base & 96.84 & 97.79 & 97.84 & 97.98 & 97.81 & \textbf{98.18} \\
& Novel & 94.00 & 93.65 & 94.29 & 94.38 & 93.88 & \textbf{95.06} \\
& HM & 95.40 & 95.68 & 96.03 & 96.15 & 95.80 & \textbf{96.59} \\

\midrule
\multirow{3}{1.5cm}{OxfordPets} & Base & 91.17 & 95.11 & 95.48 & \textbf{95.78} & 95.69 & {95.11} \\
& Novel & 97.26 & 96.57 & 97.52 & \textbf{97.65} & 97.42 & {97.27} \\
& HM & 94.12 & 95.83 & 96.49 & \textbf{96.71} & {96.55} & {96.18} \\

\midrule
\multirow{3}{1.5cm}{Stanford Cars} & Base & 63.37 & 70.72 & 75.18 & 75.75 & 80.16 & \textbf{80.53} \\
& Novel & 74.89 & 72.78 & 72.73 & 73.02 & 74.52 & \textbf{75.08} \\
& HM & 68.65 & 71.74 & 73.93 & 74.36 & 77.24 & \textbf{77.71} \\

\midrule
\multirow{3}{1.5cm}{Flowers102} & Base & 72.08 & 91.60 & 96.45 & 96.26 & 96.96 & \textbf{97.47} \\
& Novel & \textbf{77.80} & 69.62 & 74.69 & 72.62 & 76.73 & {77.58} \\
& HM & 74.83 & 79.11 & 84.19 & 82.79 & 85.67 & \textbf{86.40} \\

\midrule
\multirow{3}{1.5cm}{Food101} & Base & 90.10 & 90.17 & 90.30 & 90.36 & \textbf{90.60} & {89.60} \\
& Novel & 91.22 & 90.94 & 91.42 & 91.58 & 91.38 & \textbf{92.06} \\
& HM & 90.66 & 90.55 & 90.86 & 90.97 & \textbf{90.99} & {90.81} \\

\midrule
\multirow{3}{1.5cm}{FGVC Aircraft} & Base & 27.19 & 34.70 & 37.86 & 38.76 & \textbf{43.67} & {42.69} \\
& Novel & 36.29 & 33.53 & 34.17 & 35.08 & \textbf{36.42} & {35.21} \\
& HM & 31.09 & 34.10 & 35.92 & 36.83 & \textbf{39.72} & {38.59} \\

\midrule
\multirow{3}{1.5cm}{SUN397} & Base & 69.36 & 79.09 & 81.70 & 81.57 & \textbf{82.94} & 81.17 \\
& Novel & 75.35 & 76.85 & 77.62 & 77.92 & 78.37 & \textbf{78.98} \\
& HM & 72.23 & 77.95 & 79.61 & 79.70 & \textbf{80.59} & {80.06} \\

\midrule
\multirow{3}{1.5cm}{DTD} & Base & 53.24 & 78.67 & 79.81 & 80.81 & 82.21 & \textbf{82.45} \\
& Novel & 59.90 & 53.78 & 55.32 & 55.64 & 59.58 & \textbf{63.80} \\
& HM & 56.37 & 63.89 & 65.35 & 65.90 & 69.09 & \textbf{71.94} \\

\midrule
\multirow{3}{1.5cm}{EuroSAT} & Base & 56.48 & \textbf{94.17} & 86.98 & 92.91 & {93.06} & {92.83} \\
& Novel & 64.05 & 73.26 & 69.16 & 71.19 & 71.60 & \textbf{79.89} \\
& HM & 60.03 & 82.41 & 77.05 & 80.61 & 80.93 & \textbf{85.88} \\

\midrule
\multirow{3}{1.5cm}{UCF101} & Base & 70.53 & 83.10 & 84.38 & 84.92 & \textbf{87.05} & {86.74} \\
& Novel & 77.50 & 74.52 & 76.54 & 76.75 & \textbf{78.96} & {78.37} \\
& HM & 73.85 & 78.58 & 80.27 & 80.63 & \textbf{82.81} & {82.34} \\

\bottomrule[0.1em]
\end{tabular}}}
\label{table:more_baselines}
\vspace{-10pt}
\end{table}
Table~\ref{table:full_shallow_prompt} presents the detailed results for each dataset using manual prompts, which are summarized in Table~\ref{tab:shallow_prompt}. The manual prompts, designed for each dataset as described in~\cite{gao2023clip, zhang2022tip}, appear to underperform compared to other methods. This suggests that they may not be the optimal choice for every dataset, and that designing these prompts manually is challenging. In contrast, just ensembling multiple manual prompts~\cite{radford2021learning} works significantly better, indicating that optimal prompts may exist among these manual options. This finding also implies that utilizing improved manual prompts can substantially enhance performance, potentially replacing shallow prompts. Shallow prompt tuning for the text input yields the best results, demonstrating its effectiveness and flexibility. Therefore, we adopt this approach for our main results.

\subsection{Baseline Results with SGD}
\begin{table*}[tp]
\centering
\caption{Results for additional ablation study on configurations when combined with adaptive ensemble.}
\resizebox{\textwidth}{!}{
\setlength{\tabcolsep}{1.8pt}{
\begin{tabular}{ll|cccccccccccc}
\toprule[0.1em]
& & \begin{tabular}[c]{@{}c@{}}\textbf{Average on}\\[-0.5ex]
\textbf{11 datasets} \end{tabular} & \textbf{ImageNet} & \textbf{Caltech101}  & \textbf{OxfordPets} &  \begin{tabular}[c]{@{}c@{}}\textbf{Stanford}\\[-0.5ex]
\textbf{Cars} \end{tabular}  & \textbf{Flowers102} & \textbf{Food101} & \begin{tabular}[c]{@{}c@{}}\textbf{FGVC}\\[-0.5ex]
\textbf{Aircraft} \end{tabular} & \textbf{SUN397} & \textbf{DTD} & \textbf{EuroSAT}  & \textbf{UCF101} \\
\midrule
\multirow{3}{*}{\textbf{TPT + VA}} & Base & 83.51  & 76.43 & 98.00 & 94.76 & 79.68 & 97.28 & 89.24 & 42.27 & 80.96 & 81.49 & 92.27 & 86.24 \\
& Novel & 75.88 & 69.43 & 94.49 & 97.21 & \textbf{75.77} & 77.50 & 91.50 & 34.85 & 78.20 & 62.05 & 76.77 & 76.90 \\
& HM & 79.32 & 72.76 & 96.21 & 95.97 & 77.68 & 86.27 & 90.36 & 38.20 & 79.56 & 70.45 & 83.81 & 81.30 \\
\midrule

\multirow{3}{*}{\textbf{VPT + TA + TPT}} & Base & 83.56 & 76.93 & 98.03 & 94.77 & 79.45 & \textbf{97.51} & 89.26 & 42.14 & 81.02 & 81.72 & 92.11 & 86.21 \\
& Novel & 75.09   & \textbf{71.30} & 94.72 & \textbf{97.76} & 72.98 & 76.70 & 91.94 & 33.80 & 78.08 & 58.82 & 73.11 & 76.80\\
& HM & 78.85 & \textbf{74.01} & 96.35 & \textbf{96.24} & 76.08 & 85.86 & 90.58 & 37.51 & 79.52 & 68.40 & 81.52 & 81.23 \\
\midrule

\multirow{3}{*}{\textbf{VPT + TA (\alg)}} & Base & \textbf{83.99} & \textbf{77.12} & \textbf{98.18} & \textbf{95.11} & \textbf{80.53} & 97.47 & \textbf{89.60} & \textbf{42.69} & \textbf{81.17} & \textbf{82.45} & \textbf{92.83} & \textbf{86.74} \\
& Novel & \textbf{76.76} & {71.10} & \textbf{95.06} & 97.27 & 75.08 & \textbf{77.58} & \textbf{92.06} & \textbf{35.21} & \textbf{78.98} & \textbf{63.80} & \textbf{79.89} & \textbf{78.37}  \\
& HM & \textbf{80.04} & 73.99 & \textbf{96.59} & 96.18 & \textbf{77.71} & \textbf{86.40} & \textbf{90.81} & \textbf{38.59} & \textbf{80.06} & \textbf{71.94} & \textbf{85.88} & \textbf{82.34} \\
\bottomrule[0.1em]
\end{tabular}}}
\label{table:configuration}
\end{table*}

\begin{table*}[t]
\centering
\caption{Results for additional ablation study on scaling factor $\beta$. Our proposed methods shows robust performance on the selection of $\beta$.}
\resizebox{\textwidth}{!}{
\setlength{\tabcolsep}{1.8pt}{
\begin{tabular}{l|cccccccccccc}
\toprule[0.1em]
$\beta$ &  \begin{tabular}[c]{@{}c@{}}\textbf{Average on}\\[-0.5ex]
\textbf{11 datasets} \end{tabular} & \textbf{ImageNet} & \textbf{Caltech101}  & \textbf{OxfordPets} &  \begin{tabular}[c]{@{}c@{}}\textbf{Stanford}\\[-0.5ex]
\textbf{Cars} \end{tabular}  & \textbf{Flowers102} & \textbf{Food101} & \begin{tabular}[c]{@{}c@{}}\textbf{FGVC}\\[-0.5ex]
\textbf{Aircraft} \end{tabular} & \textbf{SUN397} & \textbf{DTD} & \textbf{EuroSAT}  & \textbf{UCF101} \\
\midrule

\textbf{$1.0$}  & 75.97 & 70.62 & 95.15 & 97.43 & 72.15 & 75.95 & 91.38 & 35.07 & 77.02 & 63.90 & 78.36 & 78.66 \\
\textbf{$2.0$}  & 76.51 & 71.06 & 95.14 & \textbf{97.44} & 73.95 & 77.06 & 91.70 & 35.35 & 78.12 & 63.99 & 78.89 & \textbf{78.92} \\
\textbf{$3.0$}  & 76.75 & \textbf{71.18} & 95.15 & 97.37 & 74.69 & 77.61 & 91.92 & \textbf{35.46} & 78.66 & \textbf{64.17} & 79.35 & 78.64 \\
\textbf{$4.0$~(\alg)} & \textbf{76.76} & 71.10 & 95.06 & 97.27 & 75.08 & 77.58 & \textbf{92.06} & 35.21 & \textbf{78.98} & 63.80 & 79.89 & 78.37 \\
\textbf{$5.0$}  & 76.72 & 71.00 & \textbf{95.16} & 97.18 & 75.10 & 77.79 & 91.96 & 35.05 & 78.96 & 63.77 & 79.88 & 78.07 \\
\textbf{$6.0$}  & 76.66 & 70.96 & \textbf{95.16} & 97.15 & \textbf{75.17} & \textbf{77.80} & 91.98 & 34.84 & 78.92 & 63.54 & \textbf{80.01} & 77.75 \\

\bottomrule[0.1em]
\end{tabular}}}
\label{table:beta_ablation}
\end{table*}

\begin{table*}[tp]
\centering
\caption{Extended results for ablation study on hyperparamter $\alpha$ related to Table~\ref{tab:motif_ensemble}.}
\resizebox{\textwidth}{!}{
\setlength{\tabcolsep}{1.8pt}{
\begin{tabular}{p{0.8cm}|cccccccccccc}
\toprule[0.1em]
$\alpha$ &  \begin{tabular}[c]{@{}c@{}}\textbf{Average on}\\[-0.5ex]
\textbf{11 datasets} \end{tabular} & \textbf{ImageNet} & \textbf{Caltech101}  & \textbf{OxfordPets} &  \begin{tabular}[c]{@{}c@{}}\textbf{Stanford}\\[-0.5ex]
\textbf{Cars} \end{tabular}  & \textbf{Flowers102} & \textbf{Food101} & \begin{tabular}[c]{@{}c@{}}\textbf{FGVC}\\[-0.5ex]
\textbf{Aircraft} \end{tabular} & \textbf{SUN397} & \textbf{DTD} & \textbf{EuroSAT}  & \textbf{UCF101} \\
\midrule

\textbf{$0.0$}  & 75.38 & 70.80 & 95.13 & 97.03 & \textbf{75.19} & \textbf{77.87} & 91.94 & 33.57 & 78.32 & 61.68 & 70.90 & 76.80 \\
\textbf{$0.1$}  & 75.86 & 71.06 & \textbf{95.19} & 97.19 & 75.17 & 77.67 & \textbf{92.10} & 34.34 & 78.82 & 62.68 & 72.74 & 77.52 \\
\textbf{$0.2$}  & 76.10 & \textbf{71.20} & 95.14 & 97.29 & 75.04 & 77.52 & 91.96 & 34.75 & 78.80 & 63.18 & 74.10 & 78.08 \\
\textbf{$0.3$}  & 76.27 & \textbf{71.20} & 95.09 & 97.39 & 74.67 & 77.33 & 91.92 & 35.16 & \textbf{78.90} & 63.74 & 75.08 & 78.54 \\
\textbf{$0.4$}  & \textbf{76.34} & 71.18 & 95.14 & 97.47 & 74.22 & 76.96 & 91.88 & 35.34 & 78.68 & 64.16 & 75.87 & 78.84 \\
\textbf{$0.5$}  & 76.29 & 71.04 & 95.15 & \textbf{97.50} & 73.59 & 76.56 & 91.78 & \textbf{35.45} & 78.40 & 64.32 & 76.41 & \textbf{79.01} \\
\textbf{$0.6$}  & 76.13 & 70.82 & 95.14 & 97.47 & 72.74 & 76.13 & 91.64 & 35.33 & 78.00 & \textbf{64.30} & 76.95 & 78.96 \\
\textbf{$0.7$}  & 75.88 & 70.46 & 95.17 & 97.39 & 71.82 & 75.66 & 91.44 & 35.25 & 77.38 & 64.23 & 77.10 & 78.79 \\
\textbf{$0.8$}  & 75.54 & 70.06 & 95.07 & 97.36 & 70.85 & 75.09 & 91.22 & 34.93 & 76.56 & 64.04 & 77.33 & 78.39 \\
\textbf{$0.9$}  & 75.10 & 69.62 & 95.01 & 97.31 & 69.63 & 74.49 & 90.98 & 34.53 & 75.68 & 63.53 & 77.44 & 77.92 \\
\textbf{$1.0$}  & 74.61 & 69.08 & 94.91 & 97.24 & 68.40 & 73.71 & 90.70 & 33.97 & 74.52 & 63.05 & \textbf{77.73} & 77.39 \\

\bottomrule[0.1em]
\end{tabular}}}
\label{table:alpha_ablation}
\end{table*}

Table~\ref{table:full_sgd} displays the reproduced results using the SGD optimizer, in contrast to the Adadelta optimizer presented in Table~\ref{table:base_to_new}. As observed, the results with SGD are slightly lower compared to those with Adadelta. This difference is likely due to the adaptive learning rate of Adadelta, which facilitates training in this unstable few-shot scenario. Nonetheless, even with the SGD optimizer, our method significantly outperforms all baselines, particularly in domains with high RTD, maintaining the same trend observed with the Adadelta optimizer.

\subsection{Comparison with More Baselines}

Due to the page limit, we present a comparison with additional baselines for base-to-novel generalization experiments in Table~\ref{table:more_baselines}, which are not included in Table~\ref{table:base_to_new}. These include training with VPT, TPT, and a combination of VPT and TPT. We also compare our method with the recently proposed PromptSRC~\cite{khattak2023self}, which employs various regularization techniques such as self-consistency loss and Gaussian averaging. Our method outperforms all these baselines in terms of harmonic mean and demonstrates particularly high performance for novel classes. Compared to PromptSRC, our method significantly outperforms in novel classes of high RTD domains, such as EuroSAT (+8.39\%) and DTD (+4.22\%), while maintaining comparable performance in other domains. Notably, our method achieves these results with a simpler training approach, without the need for numerous manual prompts for SRC loss, and with fewer hyperparameters, unlike the many required by PromptSRC’s regularization techniques. Additionally, our method surpasses the simpler baselines of naive training using VPT, TPT, and their combination, highlighting the effectiveness of our configuration design and adaptive ensemble.

\subsection{Ablation on Configuration}

To further analyze the optimal configuration in combination with an adaptive ensemble, we conduct additional ablation studies on configurations. The results, present in Table~\ref{table:configuration}, show that utilizing VPT and TA yields the best outcomes, confirming their effectiveness when paired with the adaptive ensemble. However, adding TPT to VPT and TA does not enhance performance, especially in high RTD scenarios, as evidenced by decreased performance in DTD (-4.98\%) and EuroSAT (-6.78\%) compared to configurations without TPT. While combining TPT with VA demonstrates reasonable performance, it is not as effective as the combination of VPT and TA. This highlights the importance of class separability of visual features achieved through multiple stacks of prompts. Overall, the configuration of \alg outperforms the other setups.

\subsection{Ablation on $\beta$}
\label{subsec:ablation_beta}
Table~\ref{table:beta_ablation} presents the results of an ablation study on the hyperparameter $\beta$, which is used to calculate $\alpha_{\text{eval}}$. A higher $\beta$ leads to a lower $\alpha_{\text{eval}}$, indicating greater reliance on the general knowledge of VLMs, which is beneficial for domains with low RTD, and vice versa. As observed, the performance in domains with low RTD, such as Stanford Cars and SUN397, tends to improve with a higher $\beta$. However, the optimal performance for difficult domains like Aircraft and DTD is achieved with $\beta$ values between 1.0 and 3.0. Not all domains follow this tendency since $\alpha_{\text{eval}}$ is calculated on a class-wise basis, as demonstrated in the case of EuroSAT. Interestingly, except for the value of 2.0, our method demonstrates robustness to variations in $\beta$, as it does not significantly affect the average performance. Overall, setting $\beta$ to 4.0 yields the best performance, and therefore, this value has been selected for the final results.

\subsection{Ablation on $\alpha$}
Table~\ref{table:alpha_ablation} presents the comprehensive results of the ablation study on a fixed $\alpha$, which is used in Table~\ref{tab:motif_ensemble} and Eq.~(\ref{eq:alpha}). The same $\alpha$ is applied uniformly across all classes and is set as a fixed value for both the visual and text encoders. This is done to determine the correlation between $\alpha$ and the domain, along with its transfer difficulty. Similar to Section~\ref{subsec:ablation_beta}, domains with high RTD, such as EuroSAT, require a higher $\alpha$ value to perform well compared to domains with low RTD, like Stanford Cars. These findings support the necessity for an adaptive ensemble that is closely aligned with RTD. 

\subsection{Shallow Prompt}
\definecolor{myred}{RGB}{204, 4, 67}
\definecolor{myblue}{RGB}{0,0,205}

\begin{table}[t]
\centering
\caption{Comparison of the accuracy of the base, novel, and their harmonic means among the various prompt types on text encoder.
}
\resizebox{\columnwidth}{!}{
\addtolength{\tabcolsep}{-1pt}
\begin{tabular}{l|cc|c}
\toprule[0.1em]
Prompt & Base Acc. & Novel Acc. & HM \\ \midrule
Opt. manual prompt~\cite{zhang2022tip} & \textbf{84.15} & 75.24 & 79.17 \\
Ens. (60 manual prompts~\cite{radford2021learning}) & 84.02 & 76.17 & 79.70 \\
Shallow prompt & 83.99 & \textbf{76.76} & \textbf{80.04} \\
\bottomrule[0.1em]
\end{tabular}
}\label{tab:shallow_prompt}
\vspace{-10pt}
\end{table}


Although we observe that TPT leads to overfitting, we employ one-layer learnable text prompts to enhance real-world practicality. Table~\ref{tab:shallow_prompt} compares the performance of manually optimized prompts~\cite{gao2023clip, zhang2022tip}, the ensemble of manual prompts \cite{radford2021learning}, and shallow prompts. The shallow prompt method outperforms manual prompts, proving its effectiveness. However, manual prompts, particularly when ensembled, also show comparable performance to shallow prompts, suggesting that well-designed manual prompts can be an effective alternative.

\subsection{Results on Different VLMs}
\begin{table}[t]
\centering
\tiny
\caption{Accuracy on base-to-novel generalization of \alg on EVA-CLIP~\cite{sun2023eva} and CoCa~\cite{yu2022coca}. 
}
\resizebox{1.0\columnwidth}{!}{
\setlength{\tabcolsep}{1.8pt}{
\begin{tabular}{p{1.5cm}l|ccc|ccc}
\toprule[0.1em]
\textbf{Model} & & \multicolumn{3}{c|}{EVA-CLIP-B/16} &  \multicolumn{3}{c}{CoCa-B/32} \\ \midrule
\textbf{Dataset}
& & \textbf{ZS} & {\begin{tabular}[c]{@{}c@{}}\textbf{TPT}\\[-1ex]
\textbf{+VPT} \end{tabular}} & \textbf{\alg} & \textbf{ZS} & {\begin{tabular}[c]{@{}c@{}}\textbf{TPT}\\[-1ex]
\textbf{+VPT} \end{tabular}} & \textbf{\alg} \\
\midrule
\multirow{3}{1.5cm}{Average on 11 datasets} & Base & 75.28 & 85.91 & \textbf{85.93} & 70.85 & \textbf{82.39} & 82.09 \\
& Novel & 77.68 & 75.24 & \textbf{79.34} & \textbf{74.29} & 71.05 & 73.98 \\
& HM & 76.46 & 80.22 & \textbf{82.50} & 72.53 & 76.30 & \textbf{77.87} \\

\midrule
\multirow{3}{1.5cm}{ImageNet} & Base & 79.20 & \textbf{81.78} & 81.26 & 67.10 & \textbf{69.50} & 69.46 \\
& Novel & 75.60 & 72.28 & \textbf{75.83} & \textbf{66.60} & 62.33 & 66.46 \\
& HM & 77.36 & 76.74 & \textbf{78.45} & 66.85 & 65.72 & \textbf{67.90} \\

\midrule
\multirow{3}{1.5cm}{Caltech101} & Base & 98.60 & \textbf{98.87} & 98.82 & 96.70 & 97.86 & \textbf{98.04} \\
& Novel & \textbf{97.30} & 95.05 & 97.22 & \textbf{96.30} & 94.12 & 95.98 \\
& HM & 97.95 & 96.92 & \textbf{98.01} & 96.50 & 95.95 & \textbf{97.00} \\

\midrule
\multirow{3}{1.5cm}{OxfordPets} & Base & 94.90 & \textbf{95.52} & 95.27 & 92.30 & 91.83 & \textbf{92.44} \\
& Novel & 98.10 & \textbf{98.34} & 97.97 & \textbf{96.20} & 95.07 & 93.54 \\
& HM & 96.47 & \textbf{96.91} & 96.60 & \textbf{94.21} & 93.42 & 92.99 \\

\midrule
\multirow{3}{1.5cm}{Stanford Cars} & Base & 76.90 & 85.76 & \textbf{86.16} & 84.00 & \textbf{88.94} & 88.87 \\
& Novel & \textbf{87.10} & 82.49 & 86.75 & \textbf{93.00} & 90.73 & 92.57 \\
& HM & 81.68 & 84.09 & \textbf{86.45} & 88.27 & 89.83 & \textbf{90.68} \\

\midrule
\multirow{3}{1.5cm}{Flowers102} & Base & 74.20 & 99.41 & \textbf{99.50} & 69.10 & 96.33 & \textbf{96.83} \\
& Novel & \textbf{81.10} & 77.32 & 79.94 & \textbf{74.70} & 65.61 & 70.09 \\
& HM & 77.50 & 86.98 & \textbf{88.65} & 71.79 & 78.06 & \textbf{81.32} \\

\midrule
\multirow{3}{1.5cm}{Food101} & Base & 90.30 & \textbf{90.34} & 90.24 & \textbf{81.20} & 79.87 & 80.80 \\
& Novel & \textbf{91.90} & 90.11 & 91.76 & \textbf{82.90} & 79.30 & 82.66 \\
& HM & \textbf{91.09} & 90.22 & 90.99 & \textbf{82.04} & 79.58 & 81.72 \\

\midrule
\multirow{3}{1.5cm}{FGVC Aircraft} & Base & 28.70 & 45.52 & \textbf{46.01} & 21.40 & \textbf{40.71} & 39.81 \\
& Novel & \textbf{32.50} & 26.75 & 32.12 & \textbf{25.50} & 22.04 & 25.22 \\
& HM & 30.48 & 33.70 & \textbf{37.83} & 23.27 & 28.60 & \textbf{30.88} \\

\midrule
\multirow{3}{1.5cm}{SUN397} & Base & 76.70 & \textbf{83.10} & 82.44 & 73.70 & \textbf{78.68} & 77.68 \\
& Novel & \textbf{80.80} & 76.76 & 80.54 & \textbf{77.40} & 73.50 & 77.12 \\
& HM & 78.70 & 79.80 & \textbf{81.48} & 75.50 & 76.00 & \textbf{77.40} \\

\midrule
\multirow{3}{1.5cm}{DTD} & Base & 62.80 & 83.78 & \textbf{84.15} & 62.60 & 83.04 & \textbf{83.25} \\
& Novel & 63.90 & 61.32 & \textbf{64.39} & 61.10 & 58.46 & \textbf{61.14} \\
& HM & 63.35 & 70.81 & \textbf{72.96} & 61.84 & 68.62 & \textbf{70.50} \\

\midrule
\multirow{3}{1.5cm}{EuroSAT} & Base & 72.30 & \textbf{95.32} & 94.81 & 62.80 & \textbf{96.42} & 93.87 \\
& Novel & 68.30 & 73.74 & \textbf{86.76} & 71.50 & 73.90 & \textbf{80.35} \\
& HM & 70.24 & 83.15 & \textbf{90.61} & 66.87 & 83.67 & \textbf{86.59} \\

\midrule
\multirow{3}{1.5cm}{UCF101} & Base & 73.50 & 85.58 & \textbf{86.58} & 68.50 & \textbf{83.13} & 82.01 \\
& Novel & 77.90 & 73.43 & \textbf{79.49} & \textbf{72.00} & 66.54 & 69.69 \\
& HM & 75.64 & 79.04 & \textbf{82.88} & 70.21 & 73.92 & \textbf{74.76} \\

\bottomrule[0.1em]
\end{tabular}}}
\label{table:eva_clip}
\vspace{-10pt}
\end{table}

We validate our approach using different backbones: EVA-CLIP~\cite{sun2023eva} and CoCa~\cite{yu2022coca}. Table~\ref{table:eva_clip} displays the results using these two backbones, where we compare our method with both zero-shot and naive prompt tuning approaches that combine VPT and TPT. As observed, \alg consistently outperforms the average results in terms of harmonic mean, regardless of the model used. Specifically, with EVA-CLIP, our method demonstrates superior performance for both base and novel classes. In the case of the most challenging domain, EuroSAT, our method significantly enhances performance compared to the zero-shot accuracy for novel classes (+18.46\%). A similar improvement of 8.85\% on EuroSAT is observed with CoCa. However, in terms of novel classes, the average performance of zero-shot tuning is superior for CoCa. This could be attributed to the larger patch size of this backbone, which might increase the risk of overfitting on the vision side when setting two learnable prompts. Nonetheless, our method shows comparable performance on novel classes to zero-shot CoCa, with a significant improvement in base classes. This results in superior performance in harmonic mean, demonstrating our method's effectiveness across various VLMs.



\section{Details about Observation}
\label{appendix:observation_details}

\subsection{Relative Transfer Difficulty}
\begin{table}[ht]
\centering
\caption{
    The relative transfer difficulty values for all datasets by using Definition~1.
}
\resizebox{\columnwidth}{!}{
\begin{tabular}{l|cccc}
\toprule[0.1em]
\textbf{Dataset}                     & \textbf{ImageNet} & \textbf{Caltech} & \textbf{Pets} & \textbf{Cars} \\ \midrule
\textbf{RTD}                         & $1.4\times10^{-3}$ & $1.08\times10^{-2}$ & $3.01\times10^{-2}$ & $7.7\times10^{-3}$ \\ \midrule
\textbf{Dataset} & \textbf{Flowers} & \textbf{Food} & \textbf{Aircraft} & \textbf{SUN} \\ \midrule
\textbf{RTD} & $1.52\times10^{-2}$ & $1.15\times10^{-2}$ & $4.07\times10^{-2}$ & $3.8\times10^{-3}$ \\ \midrule
\textbf{Dataset}                      & \textbf{DTD}  & \textbf{EuroSAT} & \textbf{UCF} &  \\ \midrule
\textbf{RTD}  & $4.95\times10^{-3}$ & $1.84\times10^{-1}$ & $1.42\times10^{-2}$ &  \\
\bottomrule[0.1em]
\end{tabular}
}\label{tab:rtd}
\end{table}

\begin{table}[ht]
\centering
\caption{
    The averaged cosine similarity value for inter- and intra-class and their relative ratio.
}
\resizebox{\columnwidth}{!}{
\begin{tabular}{l|cccccc}
\toprule[0.1em]
\textbf{Dataset}                     & \textbf{ImageNet} & \textbf{Caltech} & \textbf{Pets} & \textbf{Cars} & \textbf{Flowers} & \textbf{Food} \\ \midrule
\textbf{Inter}                         & 0.551 & 0.672 & 0.844 & 0.564 & 0.749 & 0.754 \\
\textbf{Intra}                         & 0.925 & 0.898 & 0.910 & 0.829 & 0.924 & 0.853 \\
\midrule
\textbf{Ratio}                         & 1.680 & 1.336 & 1.078 & 1.470 & 1.234 & 1.131 \\
\midrule
\textbf{Dataset}                     & \textbf{Aircraft} & \textbf{SUN} & \textbf{DTD}  & \textbf{EuroSAT} & \textbf{UCF} &  \\ \midrule
\textbf{Inter}                         & 0.746 & 0.487 & 0.803 & 0.896 & 0.673 &  \\
\textbf{Intra}                         & 0.858 & 0.780 & 0.823 & 0.934 & 0.866 &  \\
\midrule
\textbf{Ratio}                         & 1.150 & 1.602 & 1.025 & 1.042 & 1.287 &  \\
\bottomrule[0.1em]
\end{tabular}
}\label{tab:cosine}
\end{table}
Here, we report the value of RTD which is defined in Sectionn~\ref{sec:observation} for 11 transfer datasets. We compute the RTD based on the CLIP-B/16 model. 

\subsection{Inter- and Intra-class Cosine Similarity}
In addition to presenting relative values in Figure~\ref{fig:intra_inter}, we also report the absolute values for both inter- and intra-class similarities. We observe a significant correlation between the RTD and the ratio of intra- to inter-class similarity.

\subsection{Results on 6 datasets}
\begin{figure}[ht]
    \centering
    \includegraphics[width=1.0\linewidth]{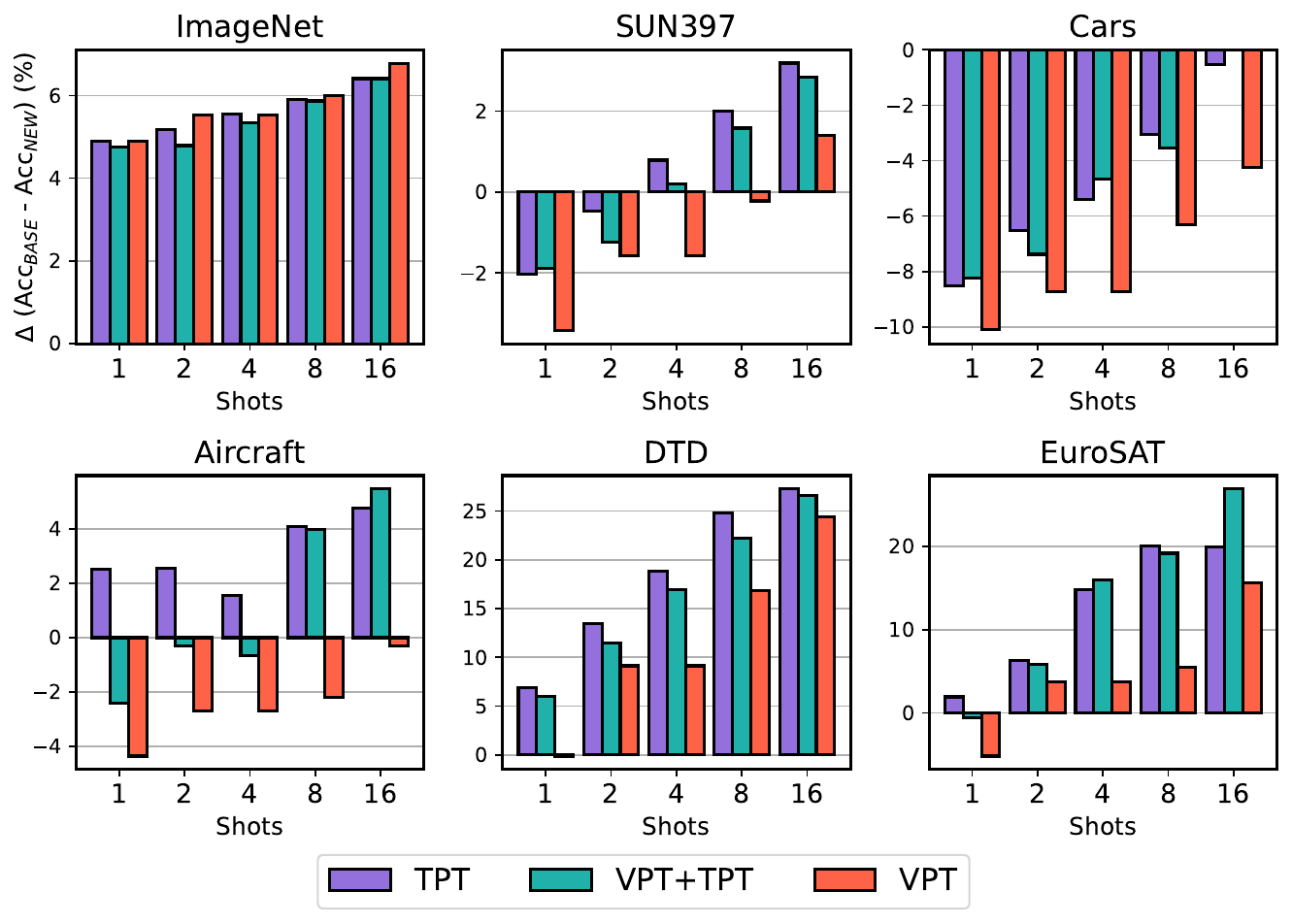}
    \caption{
    Extended results for Figure~\ref{fig:diff}. All results in different datasets show similar trends that indicate \textbf{\textcolor{vpt}{VPT}} yields a {smaller discrepancy in performance between base and novel categories}, suggesting a {reduced risk of overfitting compared to} \textbf{\textcolor{lp}{TPT}}.
    }
    \label{fig:diff2}
\end{figure}

\definecolor{lp}{RGB}{147,112,219}
\definecolor{ivlp}{RGB}{32,178,170}
\definecolor{vpt}{RGB}{255, 99, 71}

We also present extended results in Figure~\ref{fig:diff2}, which include data from three additional datasets: ImageNet, SUN397, and DTD. For ImageNet and SUN397, which already exhibit high class separability, we note that all methods—TPT, VPT, and their combination—yield similar performance differences. However, the results for DTD indicate a tendency for TPT to overfit to the base classes. This observation is consistent with the findings presented in Figure~\ref{fig:diff}.

\section{More Related Work}
\label{appendix:more_related_work}

\paragraph{Vision-Language Models}


VLMs overcome the limitations of vision-only supervised learning with their robustness and flexibility in zero-shot inference through natural language supervision. CLIP \cite{radford2021learning} facilitates this by adopting contrastive learning with a large-scale dataset of 400 million images. ALIGN\cite{jia2021scaling} further improves upon this by scaling up the dataset with more noisy image-text pairs. FILIP~\cite{yao2022filip} enables finer-grained alignment between two modalities and GLIP~\cite{li2022grounded} improves visual grounding and object detection using VLMs. CoCa~\cite{yu2022coca} employs both captioning and contrastive losses, thereby integrating the model capabilities of contrastive approaches like CLIP with those of generative methods. CyCLIP~\cite{goel2022cyclip} employs cyclic loss to ensure geometric consistency, while FLIP~\cite{li2023scaling} enhances VLMs through masking techniques. EVA-CLIP~\cite{sun2023eva} implements various training techniques, such as different attention mechanisms and optimizers, to further improve CLIP's performance. Additionally, SigLIP~\cite{zhai2023sigmoid} replaces the softmax loss with sigmoid loss, enabling more efficient pretraining with smaller batch sizes.

There is also a line of research focused on encoder-decoder or decoder-only architectures. BLIP~\cite{li2022blip} facilitates both encoding and decoding by training with three objective functions, utilizing synthetic data and data filtering. ALBEF~\cite{li2021align} employs a strategy of alignment before applying cross-attention, combined with a momentum update. Flamingo~\cite{alayrac2022flamingo} enables few-shot inference in vision-language tasks through architectural innovations, using vision-language prompts.

\paragraph{Prompt Tuning}

Efficient tuning using soft prompts, originating in the domain of natural language processing, has gained a lot of attention~\cite{lester-etal-2021-power}. This approach has also been applied in the vision-language domain to adapt to downstream tasks. CoOp~\cite{zhou2022learning} was the first to apply learnable prompts for CLIP model, replacing manual prompts for each domain. ProDA~\cite{lu2022prompt} observes that these text prompts can be viewed as a distribution and proposes prompt distributional learning for higher quality results. CoCoOp~\cite{zhou2022conditional} conditions text prompts on images to prevent overfitting to base classes. KgCoOp~\cite{yao2023visual} regularizes by minimizing the discrepancy between learned and manual prompts. UPT~\cite{zang2022unified} examines both VPT~\cite{jia2022vpt} and text prompts, proposing a unified approach to generate visual and textual prompts from the same architecture. MaPLe~\cite{khattakMaPLe} employs the alignment of visual and text prompts for improvement with deep prompts, while DCP~\cite{liu2023deeply} uses an attention mechanism for this alignment. There is also a line of research aimed at preventing the forgetting of general knowledge. ProGrad~\cite{zhu2023prompt} aligns gradient directions to preserve general knowledge, and PromptSRC~\cite{khattak2023self} utilizes multiple regularization losses with Gaussian aggregation of model weights to prevent forgetting.

\paragraph{Adapter-style Tuning}

Adapter-style tuning has been extensively explored as an alternative to prompt tuning. CLIP-Adapter~\cite{gao2023clip} was the first proposed method in this area, utilizing a two-layer MLP structure with ReLU nonlinearity in between. Additionally, it incorporates a residual connection to preserve general knowledge. For improved efficiency, Tip-Adapter~\cite{zhang2022tip} employs a cache-based model to save the features and labels of few-shot samples, using them to predict test outcomes without further training. This approach also facilitates better fine-tuning by using the cache as initial training points for further refinement. Differently, Task Residual~\cite{yu2023task} adopts a unique strategy by simply adding a residual or bias term vector for each class, reducing reliance on pre-trained features. \citet{zhu2023not} enhances cache-based models through prior refinement, which involves selecting important features for the cache-based model.



\end{document}